\documentclass[10pt,journal]{IEEEtran}
\usepackage{cite}

\usepackage{jabbrv}

\ifCLASSINFOpdf
  \usepackage[pdftex]{graphicx}
  \graphicspath{{./fig/}}
  \DeclareGraphicsExtensions{.pdf,.jpeg,.jpg,.png}
\else
\fi
\usepackage{amsmath}
\usepackage{amsthm}
\usepackage{amsfonts}
\usepackage{amssymb}
\usepackage{bbm}
\usepackage[normalem]{ulem}
\ifCLASSOPTIONcompsoc
 \usepackage[caption=false,font=normalsize,labelfont=sf,textfont=sf]{subfig}
\else
 \usepackage[caption=false,font=footnotesize]{subfig}
\fi
\usepackage{url}

\usepackage{color}
\usepackage{soul}
\soulregister\cite7
\soulregister\ref7
\soulregister\pageref7

\usepackage{multirow}
\usepackage{multicol}

\theoremstyle{definition}
\newtheorem{definition}{\textbf{Definition}}

\theoremstyle{remark}

\theoremstyle{plain}

\usepackage[linesnumbered,ruled]{algorithm2e}

\DeclareMathOperator*{\argmin}{\arg\!\min}
\usepackage{booktabs}

\usepackage[utf8]{inputenc}
\usepackage[T1]{fontenc}
\usepackage{makecell}

\newcommand{\mysemicheck}{$\checkmark\kern-1.1ex\raisebox{.7ex}{\rotatebox[origin=c]{125}{--}}$}

\begin{document}
%
\title{Concept Drift Detection via Equal Intensity $k$-means Space Partitioning}
%
%
%

\author{Anjin~Liu,~\IEEEmembership{Member,~IEEE,},~Jie~Lu,~\IEEEmembership{Fellow,~IEEE,}~and~Guangquan~Zhang
}

%
%

\markboth{IEEE TRANSACTIONS ON CYBERNETICS, VOL. XX, NO. X, SEPTEMBER 2019}
{Shell \MakeLowercase{\textit{et al.}}: Bare Demo of IEEEtran.cls for IEEE Journals}
%



\maketitle

\begin{abstract}
Data stream poses additional challenges to statistical classification tasks because distributions of the training and target samples may differ as time passes. Such distribution change in streaming data is called concept drift. Numerous histogram-based distribution change detection methods have been proposed to detect drift. Most histograms are developed on grid-based or tree-based space partitioning algorithms which makes the space partitions arbitrary, unexplainable, and may cause drift blind-spots. There is a need to improve the drift detection accuracy for histogram-based methods with the unsupervised setting.
To address this problem, we propose a cluster-based histogram, called equal intensity $k$-means space partitioning (EI-kMeans). In addition, a heuristic method to improve the sensitivity of drift detection is introduced. The fundamental idea of improving the sensitivity is to minimize the risk of creating partitions in distribution offset regions. Pearson’s chi-square test is used as the statistical hypothesis test so that the test statistics remain independent of the sample distribution. The number of bins and their shapes, which strongly influence the ability to detect drift, are determined dynamically from the sample based on an asymptotic constraint in the chi-square test. 
Accordingly, three algorithms are developed to implement concept drift detection, including a greedy centroids initialization algorithm, a cluster amplify-shrink algorithm, and a drift detection algorithm. For drift adaptation, we recommend retraining the learner if a drift is detected.
The results of experiments on synthetic and real-world datasets demonstrate the advantages of EI-kMeans and show its efficacy in detecting concept drift.
\end{abstract}


%
\IEEEpeerreviewmaketitle

%
%
%
%






\section{Introduction}
\label{s:I}

Streaming data classification consists of a routine where a model is trained on historical data and then used to classify upcoming samples. When the labels of the new arrived samples are available, they become a part of the training data. Concept drift refers to inconsistencies in data generation at different time, which means the training data and the testing data have different distributions \cite{Boracchi:QTree,Rutkowski:DTree-Gaussian-Approximation,Minku:DDD}. Drift detection aims to identify these differences with a statistical guarantee through what is, typically, a four-step process \cite{Liu:Survey}: 1) cut data stream into chunks as training/testing sets; 2) abstract the data sets into a comparable model; 3) develop a test statistical or similarity measurement to quantify the distance between the models; and 4) design a hypothesis test to investigate the null hypothesis (most often, the null hypothesis is that there is no concept drift).

Concept drift detection is also referred to change detection test or covariate shift, which is very relevant in machine learning \cite{Xin:CBCE,Xin:WEBO,Alippi:LSDD-CDT,Boracchi:QTree,Feng:TNNLS}. Some application domains are mobile tracking systems that monitor user behaviour, intrusion detection systems that identify unusual operations and remote sensing systems that reveal false sensors. In these scenarios, the systems can inference the change of situation by comparing data distributions at different time points, where the discrepancy of the distributions is estimated, based on the observed sample sets \cite{vzliobaite2016overview}. Learning under concept drift consists of three major components: concept drift detection, concept drift understanding, and concept drift adaptation \cite{Liu:Survey}. In this paper, we are focusing on improving concept drift detection accuracy on multi-cluster data sets. Regarding to the drift adaptation process, we recommend retraining the learner if a drift is confirmed as significant.


\begin{figure*}[ht]
    \centering
    \includegraphics[scale=0.5]{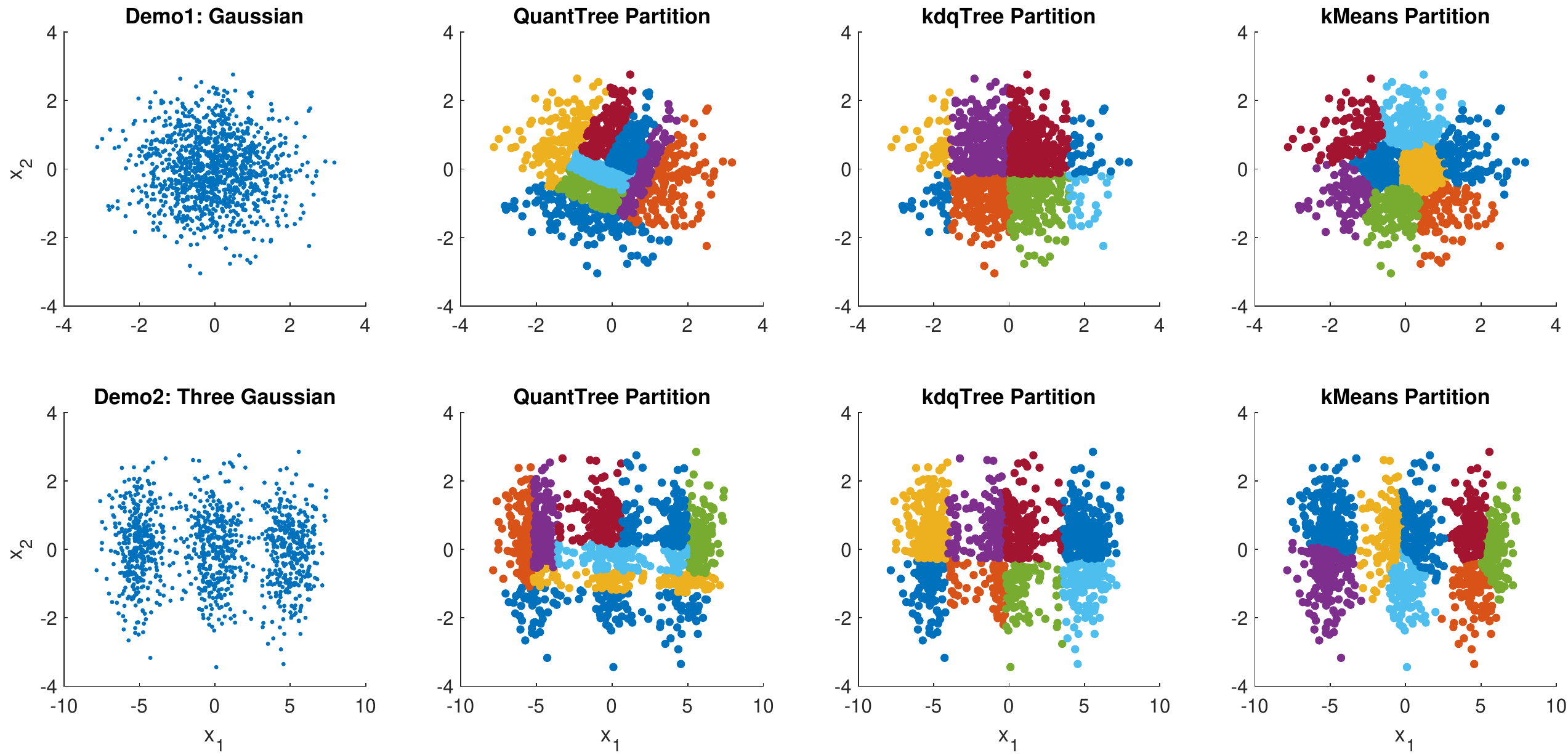}
    \caption{
    We draw a demonstration of tree-based space partitioning. Compared to cluster-based space partitioning algorithms like kMeans, tree-based space partitions are irregular and not easily understood. 
    }
    \label{fig:1}
\end{figure*}


Online and batch are two modes for drift detection \cite{Bifet:ADWIN,Alippi:LSDD-INC,Polikar:Learn2001}. Batch mode drift detection is also referred to as change-detection, or the two-sample test, where the idea is to infer whether two sample sets have been selected from the same population. This is a fundamental component of statistical data processing. 
For most change-detection algorithms, the batch size affects the drift threshold of the test statistics. Hence, extra computation are required when the batch size is not fixed \cite{Boracchi:QTree,Lu:AI1,Liu:PR}. The online approach is more flexible because the drift threshold is self-adaptive \cite{Alippi:LSDD-INC}. Alternatively, it can be calculated directly from new samples without a complicated estimation process \cite{Polikar:Learn2001}, especially when the change is simply an insertion and/or removal of observation \cite{FastKSTest}.

Histograms are the most widely used density estimators \cite{Silverman:2018}. A histogram is a set of intervals, i.e., bins, and density is then estimated by counting the number of samples in each bin. The design of the bins to reach the best density estimation result is a nontrivial problem. Most methods are based on regular grids, and the number of bins grows exponentially with the dimensionality of the data \cite{Boracchi:QTree}. A few methods instead use a tree-based partitioning scheme, which tends to scale well with high-dimensional data \cite{Boracchi:QTree,Finkel:1974}. Recent research shows that bins of equal density result in better detection performance than regular grids \cite{Boracchi:UniHist}. For example, Boracchi et al. \cite{Boracchi:QTree} developed a space partitioning algorithm, named QuantTree, that creates bins of uniform density and proved that the probabilities of these bins are independent of the data distribution. As a result, the thresholds of the test statistics calculated on these histograms can be computed numerically from uni-variate and synthetically generated data with a guaranteed false positive rate \cite{Boracchi:QTree}.

Tree-based methods have achieved outstanding results with batch mode drift detection. However, their results are less optimal with online modes due to the extra effort to recalculate the drift threshold, since their drift threshold is depend on the sample size \cite{FastKSTest}. This is a critical issue in real-world distribution change monitoring problems, particularly for streams with no explicit data batch indicators \cite{wu2016exploiting}. In addition, tree-based space partitioning does not consider the clustering properties of the data. Therefore, the partitioning results for data with complex distributions may be arbitrary, unexplainable, and may cause drift blind-spots in the leaf nodes. For example, Fig \ref{fig:1}. demonstrates the difference in the space partitioning between QuantTree, kdqTree, and kMeans algorithms. It can be seen that tree-based space partitioning will produce hyper-rectangles that crossing multiple clusters. The detected distribution change area may not be easily understood.

To address the problems caused by irregular partitions, we propose a novel space partitioning algorithm, called equal-intensity kMeans (EI-kMeans). The first priority of EI-kMeans is to build a histogram that dynamically partitions the data into an appropriate number of small clusters, then applying Pearson’s chi-square test ($\chi^2$ test) to conduct the null hypothesis test. The Pearson's chi-square test ensures the test statistics remain independent of the sample distribution and the sample size. The proposed EI-kMeans drift detection consists of three major components, which are the main contributions of this paper:
\begin{itemize}
\item A greedy equal-intensity cluster-initialization algorithm to initialize the kMeans cluster centroids. This helps the clustering algorithm to select a appropriate initialization status, and reduces the randomness of the algorithm.
\item An intensity-based cluster amplify-shrink algorithm to unify the cluster intensity ratio and ensure that each cluster has enough samples for the Pearson's chi-square test. In addition, an automatic partition number searching method that satisfies the requirements of a Pearson's chi-square test is integrated.
\item A Pearson's chi-square test-based concept drift detection algorithm that achieves higher drift sensitiveness while preserving a low false alarm rate.
\end{itemize}


The rest of this paper is organized as follows. In Section \ref{s:II}, the problem of concept drift is formulated and the preliminaries of Pearson's chi-square test are introduced. Section \ref{s:III} presents the proposed EI-kMeans space partitioning algorithm and the drift detection algorithm. Section \ref{s:IV} evaluates the space partitioning performance and the drift detection accuracy. Section \ref{s:V} concludes this study with a discussion of future work.

\section{Preliminaries and Related Works}
\label{s:II}
In this section, we define concept drift, discuss the state-of-the-art concept drift detection algorithms, and outline the preliminaries of the Pearson’s chi-square test for the proposed drift detection algorithm.

\subsection{Concept drift definitions and related works of drift detection}

Concept drift is characterized by variations in the distribution of data. In a non-stationary learning environment, the distribution of available training samples may vary with time \cite{Jose:ConceptShift,wu2016exploiting,sun2013concept, dhika:muse}. Consider a topological space feature space denoted as $X\subseteq \mathbb{R}^m$, where $m$ is the dimensionality of the feature space. A tuple $(X,y)$ denotes a data instance, where $X\in \mathcal{X}$ is the feature vector, $y\in \{y_1,\ldots,y_c \}$ is the class label and $c$ is the number of classes. A data stream can then be represented as a sequence of data instances denoted as $\mathcal{D}$. A sample set chunked from a stream via a time window strategy is a set of data instances arriving within a time interval, denoted as $D_{T_i}\in D$, where $T_i$ is the given time interval that defines the time window. A concept drift has occurred between two time windows $T_1$ and $T_2$ if the joint probability of $X$ and $y$ is different, that is, $p_{T_1}(X,y)\neq p_{T_2}(X,y)$ \cite{Gama:survey,Lu:AI2,Liu:LDD_DSDA,Alippi:HCDTs}.

According to the definition of joint probability $p(X,y)=p(y|X)\cdot p(X)$, if we only consider problems that use $X$ to infer $y$, concept drift can be divided into two sub-research topics \cite{Jose:ConceptShift,Gama:survey,Polikar:Survey,Wozniak:Survey}:
\begin{itemize}
\item Covariate shift focuses on the drift in $p(X)$ while $p(y|X)$ remains unchanged. This is considered to be virtual drift
\item Concept shift focuses on the drift in $p(y|X)$ while $p(X)$ remains unchanged. This is most commonly referred to as real drift
\end{itemize}
It is worth mentioning that $p(X)$ and $p(y|X)$ are not the only implications of $p(X,y)$ drift. The prior probabilities of classes $p(y)$ and the class conditional probabilities $p(X|y)$ may also change, which could lead to a change in $p(y|X)$ and would affect the predictions \cite{Gama:survey, dhika:Auto}. This issue is another research topic in concept drift learning that closely relates to class imbalance in data streams \cite{Xin:ImbalanceSurvey}. 

\begin{table}[h]
\caption{The capability of identifying real/virtual drift with different learning settings that are categorized by detection methods.}
\label{tab_review}
\begin{tabular}{llll}
\toprule
                               & Error rate                    & Distribution                 & Multi-hypo \\
\midrule
\multirow{4}{*}{Real drift}    & $\checkmark$ supervised       & $\checkmark\kern-1.1ex\raisebox{.7ex}{\rotatebox[origin=c]{125}{--}}$ supervised      & depends          \\
                               & $-$ unsupervised              & $\times$ unsupervised        & depends        \\
                               & $\checkmark$ semi-supervised  & $\checkmark\kern-1.1ex\raisebox{.7ex}{\rotatebox[origin=c]{125}{--}}$ semi-supervised & depends     \\
                               & $\checkmark$ active learning  & $\checkmark\kern-1.1ex\raisebox{.7ex}{\rotatebox[origin=c]{125}{--}}$ active learning & depends     \\
\midrule
\multirow{4}{*}{Virtual drift} & $\times$ supervised           & $\checkmark\kern-1.1ex\raisebox{.7ex}{\rotatebox[origin=c]{125}{--}}$ supervised      & depends          \\
                               & $-$ unsupervised              & $\checkmark\kern-1.1ex\raisebox{.7ex}{\rotatebox[origin=c]{125}{--}}$ unsupervised    & depends        \\
                               & $\times$ semi-supervised      & $\checkmark\kern-1.1ex\raisebox{.7ex}{\rotatebox[origin=c]{125}{--}}$ semi-supervised & depends     \\
                               & $\times$ active learning      & $\checkmark\kern-1.1ex\raisebox{.7ex}{\rotatebox[origin=c]{125}{--}}$ active learning & depends    \\
\bottomrule
\end{tabular}
\end{table}

Concept drift detection algorithms can be summarized in three major categories, i) error rate-based; ii) distribution-based and iii) multiple hypothesis tests (multi-hypo)\cite{Liu:Survey}. The algorithms can also be distinguished in different learning settings, such as supervised, unsupervised \cite{sethi2017reliable, Liu:TNNLS}, semi-supervised \cite{haque2016sand}, and active learning settings\cite{vzliobaite2013active}.
For a supervised setting, the target variable is available for drift detection. Most error rate-based drift detection algorithms are developed with this setting \cite{shao2017robust, barros2017rddm}. In later work, the problem of label availability in data streams with concept drift has been acknowledged\cite{widyantoro2005relevant, huang2007active} pointing out concept drift could occur within unsupervised and semi-supervised learning environments. Accordingly, active learning strategy is adopted by\cite{vzliobaite2013active} to address concept for improving the learning performance.

Real and virtual are two major drift $types$. Error-based, distribution-based and multiple hypothesis are three major $categories$ of drift detection algorithms. Supervised, unsupervised, semi-supervised and active learning are four major $settings$ of learning under concept drift. In Table \ref{tab_review}, the $\checkmark$ indicates the algorithms in this category can detect and distinguish different drift types with the given setting. The $\checkmark\kern-1.1ex\raisebox{.7ex}{\protect\rotatebox[origin=c]{125}{--}}$ indicates the they can detect drifts but cannot distinguish the types. The $\times$ indicates they are unable to detect the drift. The $-$ indicates the algorithms in this category cannot be applied in the given setting. With regard to multiple hypothesis tests, the capability of these algorithms varies significantly, since they could be a combination of multiple error-based algorithms or a hybrid of both error and distribution-based algorithms. Therefore, it is hard to give a conclusion for this category. In addition, it is worth to mention that Mahardhika\mbox{\cite{dhika:weaklysup}} has proposed a method to handle concept drift in a weakly supervised setting.

EI-kMeans is one distribution-based drift detection algorithm. Most Hoeffding bound-based algorithms, like\cite{I:HDDM, pesaranghader2016fast}, belong to error rate-based drift detection that can only detect real drift with supervised, semi-supervised or active learning settings. The main contribution of EI-kMeans is different from conventional distribution-based drift detection. Conventional distribution-based drift detection algorithms aim to find a novel test statistics to measure the discrepancy between two distributions and to design a tailored hypothesis test to determine the drift significant level, such as \cite{Liu:PR, Alippi:LSDD-INC, Kuncheva:PCA}. In contrast, EI-kMeans focuses on how to efficiently convert multivariate samples into a multinomial distribution and then use an existing hypothesis test to detect the drift. Since EI-kMeans is using Pearson's chi-square test as the hypothesis test, the drift threshold can be calculated directly according to Chi-square distribution and it can be implemented in an online manner. Other distribution-based algorithms may need to re-compute the drift threshold as new samples become available.


\subsection{Histogram-based distribution change detection}

Histograms are the oldest and most widely used density estimator \cite{Silverman:2018}. The bins of the histogram are the intervals, i.e., partitions, of the feature space. Hence, a $K$-bins histogram is a set of $K$ partitions, denoted as $\{S_k\}_{k=1,\ldots,K}$, where $S_k$ is a partition of the feature space $\mathcal{X}$, $S_k\subseteq\mathcal{X}$, $\bigcup_{k=1}^K S_k = \mathcal{X}$ and $S_i\cap S_j=\phi$, for $i\neq j$ \cite{Boracchi:QTree}. Histograms are often built upon regular grids, which means the number of bins will grow exponentially along with the dimensionality of the data \cite{Boracchi:QTree}. Dasu et al. extended QuadTree \cite{Finkel:1974} based on the idea of a k-dimensional tree \cite{Bentley:1975} and developed a kdqTree space partitioning scheme \cite{Dasu:kdqTree}. In the kdqTree scheme, the feature space is partitioned into adaptable cells of a minimum size and a minimum number of training samples. Then, the Kullback–Leibler divergence is used to quantify the distribution discrepancy, and bootstrap sampling is used to estimate the confidence interval. Another recent tree-based space partitioning algorithm, named QuantTree, was proposed by Boracchi et al. \cite{Boracchi:QTree}, which splits the feature space into partitions of uniform density. The advantages of QuantTree is that the test statistics computed based on it are distribution free \cite{Boracchi:QTree}. 

\begin{figure}
    \centering
    \includegraphics[scale=0.55]{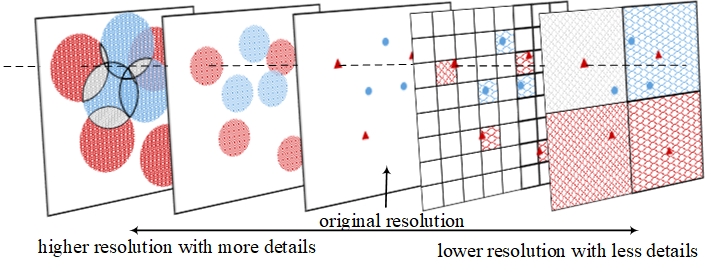}
    \caption{A demonstration of lower and higher resolution space partitioning \cite{Liu:PR}. With lower resolutions, density is estimated by counting the sample points in the partitions. With higher resolutions, density is estimated with a smoothing function, such as kernel density estimation.
    }
    \label{fig:2}
\end{figure}

Distribution change detection with histograms can be considered from the perspective of granularity and can be categorized into two groups: higher resolution histograms and lower resolution histograms, as demonstrated in Fig. \ref{fig:2}.

Lower resolution partitioning requires a large number of training samples so that each partition could have enough samples to estimate the density. Without adequate training samples, the estimate the density may suffer from randomness. To mitigate this problem, Lu et al. \cite{Lu:AI2,Lu:AI1} proposed a competence-based space partitioning method that uses related sets to enrich sample sets, then applying space partitioning and calculating the distribution discrepancy. Liu et al. applied a similar strategy \cite{Liu:PR} by partitioning the feature space based on k-nearest neighbor particles. These higher-resolution partitions resulted in higher drift detection accuracy on small sample sets, but also suffered from higher computational costs.

\subsection{Pearson’s chi-square test}

Pearson’s chi-square test, or $\chi^2$ test for short, is used to determine whether there is a significant difference between the expected frequencies and the observed frequencies in one or more sets of data \cite{box:sampleSize}. The test statistic follows a chi-square distribution when there is no significant difference. The purpose of the test is to assume the null hypothesis is true and then evaluate how likely a specific observation would be.

The standard process of the $\chi^2$ test is to use sample data to find: the degrees of freedom, the expected frequencies, the test statistic, and the $p$-value associated with the test statistic \cite{box:sampleSize}. Given a contingency table with $i$ rows and $j$ categorical variables (column), the degrees of freedom are equal to 
\begin{equation*}
    DF=(i-1)(j-1).
\end{equation*}

The expected frequency counts are computed separately for each level of one categorical variable at each level of the other categorical variable. The $i$th and $j$th expected frequencies of the contingency table are calculated with the equation 
\begin{equation*}
    E_{i,j}=(n_i\times n_j )/n,
\end{equation*}
where $n_i$ is the sum of the frequencies for all columns in row $i$, $n_j$ is the sum of the frequencies for all rows of columns $j$, and $n$ is the sum of all rows and columns. The test statistic is a chi-square random variable $\chi^2$ defined by 
\begin{equation}\label{eq:1}
    \chi^2=\sum \frac{(O_{i,j}-E_{i,j})^2}{E_{i,j}},
\end{equation}
where $O_{i,j}$ is the observed frequency count at row $i$ and column $j$, and $E_{i,j}$ is the expected frequency count at row $i$ and column $j$. The $p$-value is the probability of observing a sample statistic as extreme as the test statistic. Since the p-value is a $\chi^2$ test statistic, it can be computed with the chi-square probability distribution function.

Pearson's chi-square test should be used with the conditions described in \cite{box:sampleSize}, which assumes there is a sufficiently large sample set. If the $\chi^2$ test is applied to a small sample set, the $\chi^2$ test will yield an inaccurate inference and will result in a high Type II error. The true positive detection accuracy will be impaired, but the false alarm rate will not increase. According to the central limit theorem, an $\chi^2$ distribution is the sum of $O_{i,j}$ independent random variables with a finite mean and variance that converges to a normal distribution for large $O_{i,j}$. For many practical purposes, Box et al. \cite{box:sampleSize} claim that for $O_{i,j}>50$ and $E_{i,j}>5$ the distribution of the estimated test statistics is sufficiently close to a normal distribution for the difference to be ignored. In other words, to avoid the bias raised by asymptotic issues, the observations and expectation frequencies should be greater than a particular threshold.

\section{EI-kMeans Space Partitioning and Drift Detection with Pearson’s Chi-square Test}
\label{s:III}

This section presents our EI-kMeans space partitioning histogram and our drift detection algorithm based on Pearson’s chi-square test. The algorithm implementation detail is given, and the complexity is discussed at the end of this section. 

\subsection{The risk of offset partitions in histogram-based drift detection}
\label{s:III-A}


Let us begin by restating the concept drift detection problem and our proposition.

\textbf{Problem. 1}. Let $d_{T_1}$ and $d_{T_2}$ be random variables defined on a topological space $X\subseteq \mathbb{R}^m$, with respect to $p_{T_1},p_{T_2}\in P(\mathcal{X})$, where $P(\mathcal{X})$ consists of all Borel probability measures on $\mathcal{X}$. Given the observations $D_{T_1}=\{d_{11},\ldots,d_{1m_1}\}$ and $D_{T_2}=\{d_{21},\ldots,d_{2m_2}\}$ from $d_{T_1}$ and $d_{T_2}$, respectively, how much confidence do we have that $d_{T_1}\neq d_{T_2}$?
At present, most distribution change detection methods assume that the observations $D_{T_1 },D_{T_2}$ are i.i.d. which makes the assumption and objective equivalent to a two-sample test problem. 

The problem of analyzing a data stream to detect changes in data generating distribution is very relevant in machine-learning and is typically addressed in an unsupervised manner \cite{Boracchi:QTree}. However, it can easily be extended to handle a supervised setting. For this, there are two options for implementing our proposed solution without changing the algorithms. Option 1: Considering the label or target variable as one feature of the observations in the sample set and then applying the proposed concept drift detection algorithms.
Option 2: Separate the observations based on their labels and detect concept drift individually. 

The design of the space partitioning algorithm is critical to how the histogram is constructed, but nowhere in the literature is there a definitive conclusion on how to build a perfect histogram. Tree-based histogram construction is one of the most popular methods for change detection. QuantTree \cite{Boracchi:QTree} is a representative algorithm that creates partitions of uniform density in a tree structure. Given all the distributions are the same, the drift threshold is independent of the data samples and can be numerically computed from univariate and synthetically generated data. Although some studies claim that uniform-density partition schemes are superior based on experiment evaluations \cite{Boracchi:UniHist,Liu:PR}, no study includes a detailed justification of its claims.

The fundamental idea of drift detection via histograms is to convert the problem of a multivariate two-sample test into a goodness-of-fit test for multinomial distributions. If the data is categorical and belongs to a collection of discrete non-overlapping classes, it has a multinomial population \cite{Liu:Survey}. In this case, each partition (each bin in the histogram) constitutes a categorical, non-overlapping class. And the null hypothesis for a goodness-of-fit test to evaluate how the observed frequency $O_{i,j}$ match the expected frequency $E_{i,j}$, that is, the number of testing data in a partition is expected to fall into an estimated range based on the training data \cite{box:sampleSize}. The hypothesis is rejected if the p-value of the observed test statistic is less than a given significance level $\alpha$ \cite{Alippi:HCDTs,Alippi:JIT_RecurrentDrift,song2019fuzzy}.

Pearson's chi-square test is a commonly used hypothesis test for this task if the expected frequency for each category is larger than 5, and the observed frequency for each category is larger than 50 \cite{box:sampleSize}. In histogram-based drift detection, this requirement can be satisfied by controlling the number of samples in partitions, such as reducing the number of partitions K to ensure all partitions contain enough samples. Recall the $\chi^2$ test statistic in Eq. \eqref{eq:1}., we know that, given the same number of partitions, the higher the value of the test statistic, the more likely it is that a distribution drift has occurred. Therefore, the objective is to design a partition algorithm to have the highest $\chi^2$ test statistic. If the highest $\chi^2$ test statistic, which represents the highest distribution discrepancy, does not refute the null hypothesis, then there is no drift. Theoretically, the expected frequency counts for all partitions becomes known once the partition scheme is determined.

To maximize the $\chi^2$ test statistic, the space partitioning strategy needs to avoid partitions that have distribution discrepancies that cannot be measured by subtracting the observations and expectations, as illustrated in Fig. 3. Related defintions are given below.

\begin{figure}
    \centering
    \includegraphics[scale=0.5]{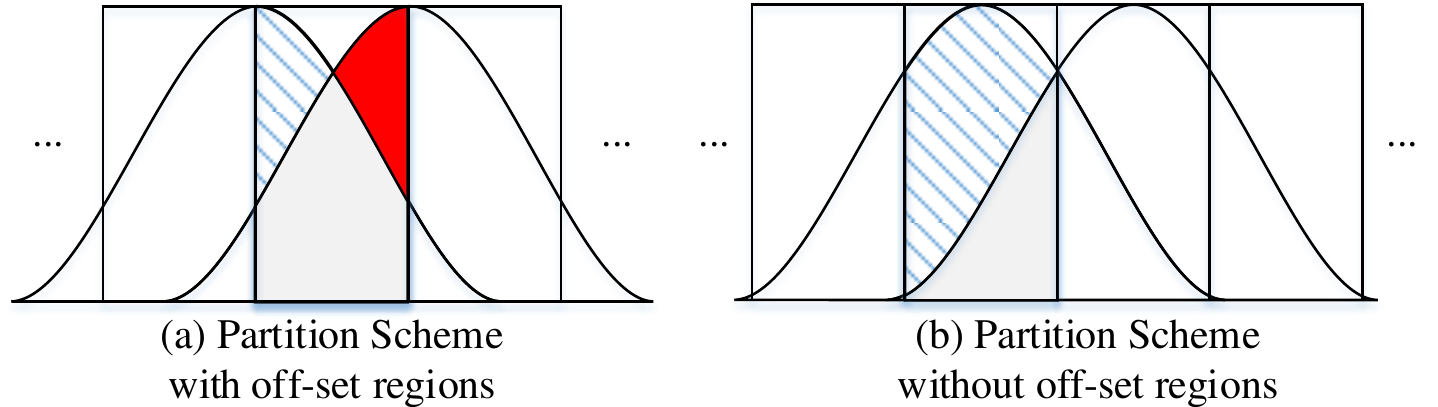}
    \caption{A demonstration of the offset partition. The toy data represents a Gaussian distribution with a mean shift. The partition scheme in (a) has an offset partition (highlighted in the middle), while the partition scheme in (b) does not. For (a), even though there is a distribution discrepancy, the value of $(O_{i,j}-E_{i,j})^2$ in the middle partition is equal to 0.
    }
    \label{fig:3}
\end{figure}

\begin{definition}(Partition Absolute Variation) The absolute variation of a partition is defined as the integration of the probability density difference of $p_{train}(x)$ and $p_{test}(x)$ in partition $S_k$, denoted as
    \begin{equation*}
        \delta_{S_k}^{av} (p_{train}(x),p_{test}(x)) = \int_{S_k}\Bigl|p_{train}(x)-p_{test} (x)\Bigr|dx.
    \end{equation*}
where $p_{train}(x)$, $p_{test}(x)$ denotes the probability density function of the training and testing data, and ${S_k}$ is the partition interval.
\end{definition}

\begin{definition}(Partition Probability Variation) The probability variation of a partition is defined as the difference of the integration of the probability density in partition $S_k$ of $p_{train}(x)$ and $p_{test}(x)$, denoted as
    \begin{equation*}
        \delta_{S_k}^{pv} (p_{train}(x),p_{test}(x)) = \Bigl|\int_{S_k}p_{train}(x)dx-\int_{S_k}p_{test} (x)dx\Bigr|
    \end{equation*}
\end{definition}
Then we have the offset partition defined as follow.
\begin{definition}(Offset Partition)
    Given two probability density distributions $p_{train}(x)$ and $p_{test}(x)$, a space partition $S_k$ is an offset partition if the absolute variation is larger than the probability variation, denoted as $\delta_{S_k}^{av} (p_{train}(x),p_{test}(x)) > \delta_{S_k}^{pv} (p_{train}(x),p_{test}(x))$.
\end{definition}

Concept drift detection requires the histogram built on training samples only. Admittedly, the impact of offset partitions on distribution estimation can be reduced by learning methods that optimize the density difference between the training and testing samples. However, this method can be time-consuming and is not feasible when the testing data is small or even not available. Additionally, this method may not be suitable for streaming data since data may arrive much faster than it can be tested. Therefore, for concept drift detection, histograms need to be designed only based on training data, and minimizing the occurrence of offset partitions. In other words, to achieve the best drift detection results, the histogram should have the least number of offset partitions. To detect concept drift, we propose the following strategies to reduce the appearance of offset partitions.

\begin{figure}
    \centering
    \includegraphics[scale=0.5]{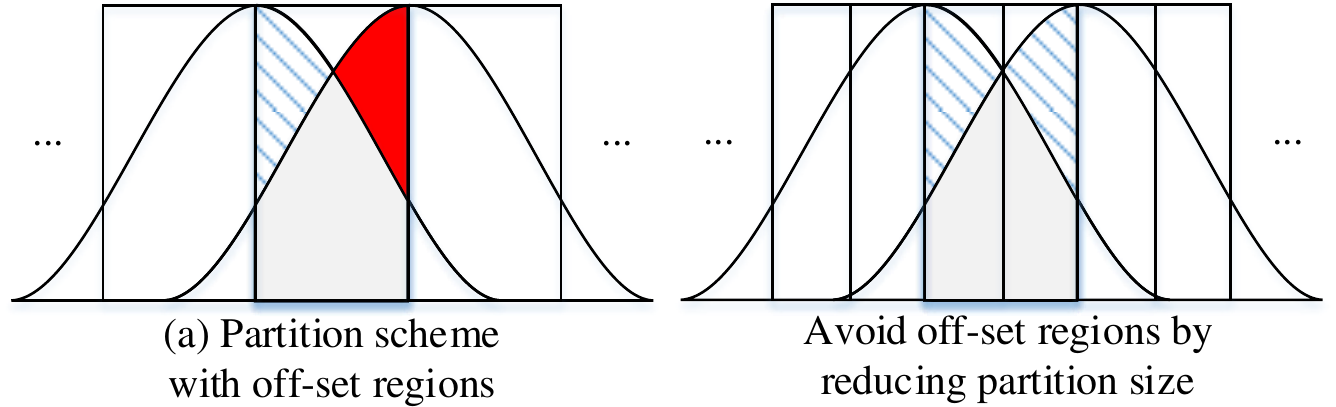}
    \caption{Reducing the risk of an offset region by reducing partition size. Recall the example in Fig. \ref{fig:3}. The risk of having partitions on an offset region, as shown in (a), can be reduced by minimizing the interval size of the partition, as shown in (b). However, offset regions cannot be completely avoided because the drift direction and margin are unknown.
    }
    \label{fig:4}
\end{figure}

\begin{itemize}
\item \textit{Partitions should avoid cluster gaps}. With multi-cluster training sample sets, there are gaps between clusters. If a partition steps across multiple clusters, its sensitivity to drift will be affected. 
\item \textit{Partitions should be as compact as possible}. The distances between samples within a partition should be minimized. If the drift direction is unknown, one rule of thumb to avoid offset partitions is to keep the shape of the partition as compact as possible, as shown in Fig. 4. However, this strategy must be constrained by a predefined minimum partition size. Otherwise, the partitions will be too small to yield statistical information. In our case, the $\chi^2$ test requires the number of observations to be as large possible. The minimum requirement is 50 observations for each partition, and the expected frequency count has to be greater or equal to 5 \cite{box:sampleSize}.
\end{itemize}

This strategy also conforms to Boracchi et al.’s \cite{Boracchi:UniHist} conclusion that histogram bins of equal density provide better detection performance than regular grids. For example, given a sample set with 1000 samples and 50 as the minimum number of points in a partition with no identical samples, the smallest average interval size of 1000/50=20 partitions is always smaller than the smallest average interval size of 19 partitions. However, histogram bins of equal density may not always have the smallest average interval size. Therefore, bins of non-uniform density may provide superior performance to uniform density bins in some cases.

The distribution discrepancy within partitions is also important, which is another issue resulting from offset partitions that may influence the drift detection results. An $\chi^2$ test cannot identify a distribution discrepancy inside a partition, so the histogram design should ensure the distributions in the partitions are as simple as possible. For example, kernel density estimation-based methods assume the data follows Gaussian mixture distributions. However, this can cause bandwidth selection problems. Therefore, we need an indicator that represents whether or not the density of samples in the same partition are similar. Definition 4 defines this indicator as the offset margin:

\begin{definition}(Offset Margin)
    The off-set margin of partition $S_k$ is defined as the difference between the absolute variation and the probability variation of $S_k$, denoted as $\Delta(S_k)=|\delta_{S_k}^{av} (p_{train},p_{test})-\delta_{S_k}^{pv} (p_{train},p_{test})|$.
\end{definition}

The offset partition is only one of many issues that might influence the detection results. Intuitively, the more partitions we have, the less likely offset partitions occur. Also, different partitioning schemes will result in different drift detection results with different sample sets. Minimizing the risk of offset partitions may result in better performance generally, but it may not be the best choice for a particular sample set.
\subsection{EI-kMeans Space Partitioning}
\label{s:III-B}

Since the main objective is to keep the risk of offset regions as small as possible without knowing the testing data, the simplest method is to create as many partitions as possible. To this end, we propose using the average partition interval size as an indicator for constructing a histogram. The $\chi^2$ test requires there be more than 50 observations with an expected frequency greater than 5. This requirement can be satisfied by adding constraints onto the indicator. The general form of the objective function is to find the centroids with the smallest average interval size:
\begin{equation}
    \argmin_{\{S_k\}_{1,\ldots,K}} \frac{1}{K}\sum_{k=1}^{K}\mathrm{Vol}(S_k), s.t. O_k \ge 50, and~E_k \ge 5.
\end{equation}
As the interval in high dimensional cases is a volume, the interval size is denoted as $\mathrm{Vol}(S_k)$. The $O_k$ denotes the count of observations in $S_k$, and $E_k$ denotes the expected frequency count in $S_k$. 

\begin{figure}
    \centering
    \includegraphics[scale=0.5]{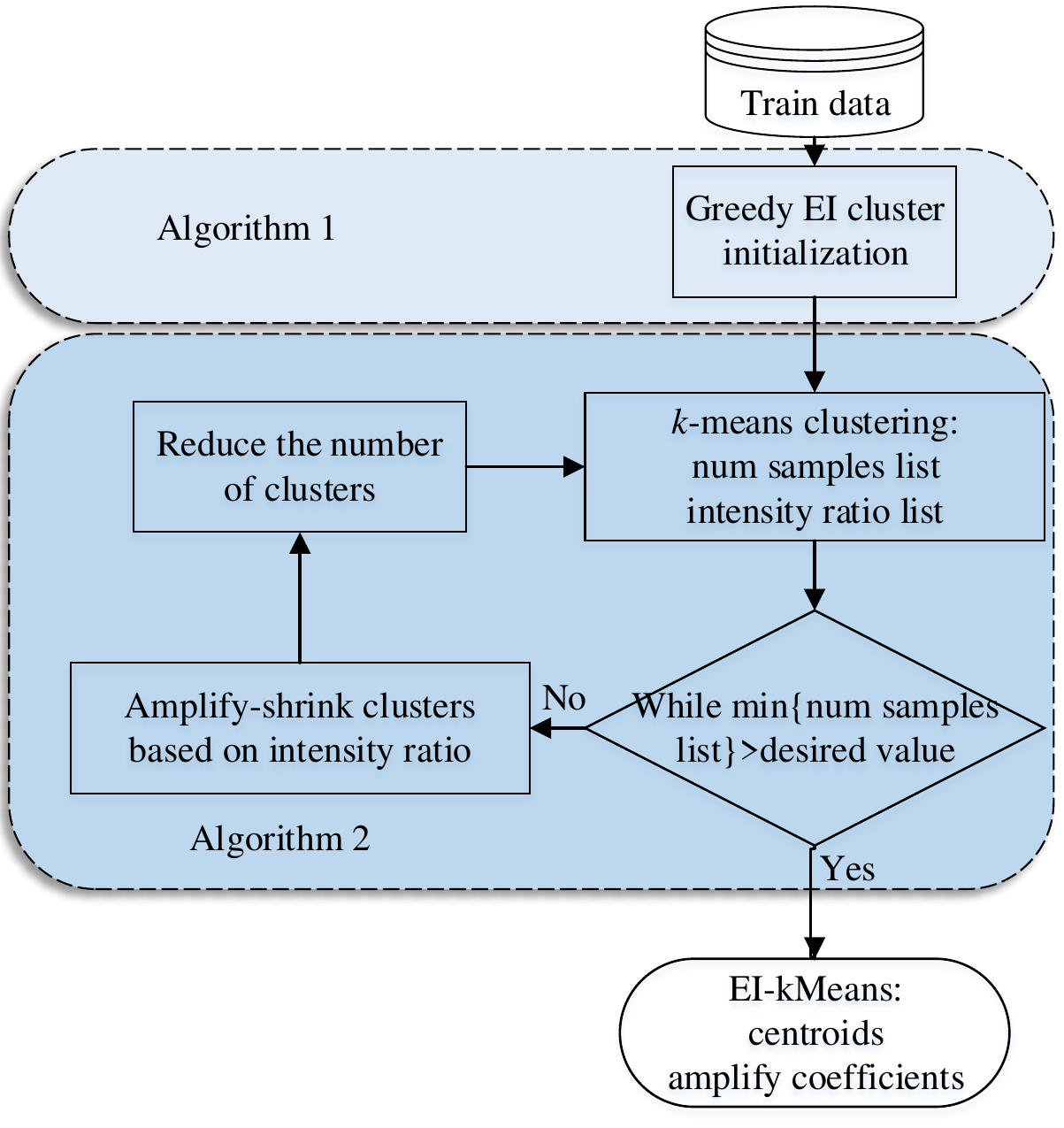}
    \caption{
    The workflow for constructing the EI-kMeans histogram.
    }
    \label{fig:5_1}
\end{figure}

The nature of kMeans makes it a good option for this task. Adding constraints can be addressed by introducing an algorithm to monitor the number of samples in each cluster. Here, the volume indicator represents the average distance to the centroids. The overall workflow of EI-kMeans space partition is shown in Fig. \ref{fig:5_1}.

As shown in Fig. \ref{fig:5_1}, the procedure begins by initializing the cluster centroids with a greedy equal-intensity k-means initialization algorithm. The objective of this algorithm is to segment the feature space into a set of partitions with the same number of samples. Let  $D$ denote the training data set for the histogram, and $D_{S_k}$  be the samples located in partition $S_k$. There are $K$ partitions. The greedy equal intensity kMeans initialization will evenly divide the samples into $K$ groups. The centroids of these groups will be input into kMeans as the initial centroids. Once thekMeans converges or reaches the maximum iteration criteria, the returned sample labels and the centroids are used for equal-intensity cluster amplification. 

Greedy equal-intensity k-means initialization finds the farthest sample, i.e., the sample with the longest distance to its nearest neighbor. The $\frac{n}{K}$-nearest neighbors of this sample is labelled as the first partition, where $n=|D|$ is the cardinality of the training sample set. The labelled samples are then removed from the training set, and the above process is repeated until all the samples are labelled.

\textbf{Remark}: if the remainder of $\frac{n}{K}$ is not equal to 0, the remainder will be evenly distributed into the first few partitions, that is, samples with $\lfloor \frac{n}{K}\rfloor+1$ nearest neighbors will be labeled instead of those with the $\frac{n}{K}$ nearest neighbours.

\begin{algorithm}[ht]
    \caption{Greedy equal-intensity kMeans centroids initialization}
    \label{alg:1}
    \small
    \SetKwInOut{Input}{input}
    \SetKwInOut{Output}{output}

    \Input{1. Training set, $D$
        \newline 2. The number of cluster, $K$
        }
    \Output{Centroids list, $\mathcal{C}=\{C_1 ,\ldots, C_k\}$}
    \BlankLine
    Initialize the expected number of samples for each partition $V_N=\{n_k\}_{k=1,\ldots,K}$\;

    \For {$k$ in range $K$} {
        
        Find the 1NN for all samples in $D$\;
        Sort the 1NN distance for all samples\;
        Get the sample with the largest value of 1NN distance\;
        Find the $n_k$ nearest neighbors of this sample\;
        Calculate the mean as the centroids, $C_k$\;
        Append the centroids to the output list, $ \mathcal{C}=\mathcal{C}\bigcup\{C_k\}$ \;
    }
    \Return $\mathcal{C}=\{C_1 ,\ldots, C_k\}$\;
\end{algorithm}

Algorithm 1 shows the pseudocode for the greedy equal-intensity kMeans centroid initialization. The inputs are: a set of training samples $D$; and the number of clusters to initialize. In this algorithm, one trick we used to control the computation cost is sub-sampling. The input training set $D$ could be the entire training set or just a subset of the training set. Some data pre-processing techniques, such as dimensionality reduction or data normalization, will be applied before running our algorithm. Since different data sets may require different data pre-processing techniques, this is not the main scope of our algorithm.

Denote the number of samples in dataset $D$ as $n$, and $n=\sum_{k=1}^K n_k$. 
The runtime complexity in line 3 is $\mathcal{O}(n\log n)$ with an appropriate nearest neighbor search algorithm, such as $k$-d tree. The sorting complexity for line 4 is $\mathcal{O}(n \log n)$ with a merge-sort algorithm. The complexity for lines 6 and 7 are $\mathcal{O}(n_k \log n)$ and $\mathcal{O}(n_k)$, respectively. Therefore, the total complexity for each iteration is $\mathcal{O}(n\log n)$ according to the rule of sums. The total complexity for the greedy equal-intensity k-means centroids initialization is $\mathcal{O}(nK\log n)$

Based on the labels, the cluster sample intensity ratio is computed by dividing the count of samples in a cluster by the total number of training samples, i.e., 
\begin{equation}
\label{eq:intensity}
    r_{S_k}=\frac{n_{S_k}}{n},
\end{equation}
where $n_{S_k}=|D_{S_k}|$ is the number of samples located in partition $S_k$.

The intensity ratios for all clusters can be represented as a vector $V_r$, where the shape of $V_r$ is $K\times 1$. The amplify coefficient function for the cluster distance is calculated based on this vector:
\begin{equation*}
    V_{coe}=e^{\theta(V_r-1)},
\end{equation*}
where $\theta$ is a parameter to control the shape of the coefficient function. To convert the amplify coefficients to matrix, the amplify coefficient vector $V_{coe}$ is multiplied by a $1\times n$ vector to create a $K\times n$ amplify coefficient matrix, denoted as $M_{coe}=V_{coe}\cdot V_1$. Calculating the paired Euclidean distance matrix between the centroids and the data samples as $M_{dist}$, the amplified distance matrix is derived by
\begin{equation*}
    \mathcal{M}_{dist}=M_{dist} M_{coe},
\end{equation*}
and the amplified cluster labels are derived by finding the centriod index with the minimum amplified distance,
\begin{equation*}
    \hat{y} =\argmin_{\{1,\ldots,K\}} \mathcal{M}_{dist}.
\end{equation*}

In the cluster amplify-shrink algorithm, $\theta$ is chosen through a grid search from a predefined set $\Theta=\{0,0.05,\ldots, 1.5\}$. When $\theta=0$, the amplify coefficients are all equal to 1, which will not amplify or shrink any of the clusters. As $\theta$ increases, the clusters are amplified or shrunk sharply. If the minimum number of samples in a partition is larger than the desired value, the amplify-shrink algorithm will terminate, denoted as
\begin{equation*}
    \min \{n_{S_k}\}_{k=1,\ldots,K} \ge \beta,
\end{equation*}
where $\beta$ is the desired value of the minimum number of samples in the partitions. According to the requirements of Pearson’s chi-square test, the desired value is $\beta=50$. If no $\theta$ can satisfy the desired value, the number of partitions is reduced by 1, namely $K=K-1$, and the above process is repeated.

\begin{algorithm}[ht]
    \caption{Equal intensity k-means space partitioning}
    \label{alg:2}
    \small
    \SetKwInOut{Input}{input}
    \SetKwInOut{Output}{output}

    \Input{1. Training set, $D$
        \newline 2. The number of cluster, $K$
        \newline 3. Minimum number of samples in partition, $\beta=50$
        \newline 4. The amplify coefficient function parameter range, $\Theta=\{0, 0.05,\ldots, 1.5\}$
        }
    \Output{EI-kMeans Histogram $(\mathcal{C},V_{coe})$:
    }
    \BlankLine
    Initialize the number of clusters $K=\frac{n}{\beta}$, where $n=|D|$\;
    Initialize $\hat{V}_N=\{0,\ldots, 0\}$\;
    \While{$\min\{\hat{V}_N\} < \beta$ and $K>1$}{
        Initialize the centroids, $\mathcal{C}_{ini}=$GreedyInitial$(D,K)$\;
        Initialize the expected number of samples for each partition $V_N$\;
        Update centroids, $\mathcal{C}=k$-means($D$, $\mathcal{C}_{ini}$)\;
        Count the samples in each partition, $\hat{V}_N$\;
        Calculate the intensity vector, $V_r=\frac{\hat{V}_N}{V_N}$\;
        
        \For {$\theta$ in range $\Theta$} {
            Calculate the amplify coefficient vector, $V_{coe}$\;
            Calculate the $D$ to $\mathcal{C}$ distance matrix, $M_{dist}$\;
            Calculate the amplified distance, $\mathcal{M}_{dist}$\;
            Get the new labels, $\hat{y}$\;
            Count the samples in each partition, $\hat{V}_N$\;
        }
        
        \If{$\min\{\hat{V}_N\} \ge \beta$}{
            return EI-kMeans Histogram $(\mathcal{C},V_{coe})$\;
        }
    }
    $\mathcal{C}=k$-means($D$, $\mathcal{C}_{rand})$\;
    $V_{coe}=\{1,\ldots, 1\}$\;
    \Return EI-kMeans Histogram $(\mathcal{C},V_{coe})$\;
\end{algorithm}

Algorithm 2 shows the pseudocode for the equal-intensity kMeans space partitioning. The inputs are: the training set $D$; the minimum number of samples in a partition $\beta$; and the grid search range of the amplify coefficient function parameter $\Theta$. The aim of lines 4-8 are to build kMeans clusters of similar intensity. Then, from lines 10 to 19, the clusters are amplified or shrunk based on their intensity ratio. The amplify-shrink process will end up satisfying the minimum number of samples or until it reaches the end of the range of $\Theta$. If a desired partition sets cannot be found after the amplify-shrink process, the number of clusters will be reduced by 1, namely $K$ is updated to $K-1$, and the process is repeated.

From lines 10 to 19, the main cost is the multiplication of the matrix, which has a runtime complexity equal to $\mathcal{O}(|\Theta|Kn)$. Because $|\Theta|$ is constant, the complexity is actually $\mathcal{O}(Kn)$. The greedy initialization in line 4 is $\mathcal{O}(Kn log n )$, the k-means in line 6 is $\mathcal{O}(Kn)$. Considering the while loop starts from K to 2, the worst-case complexity of EI-kMeans space partition is $\mathcal{O}(K^2 n log n )$.

\subsection{EI-kMeans Drift Detection}
\label{s:III-C}

\begin{figure}
    \centering
    \includegraphics[scale=0.5]{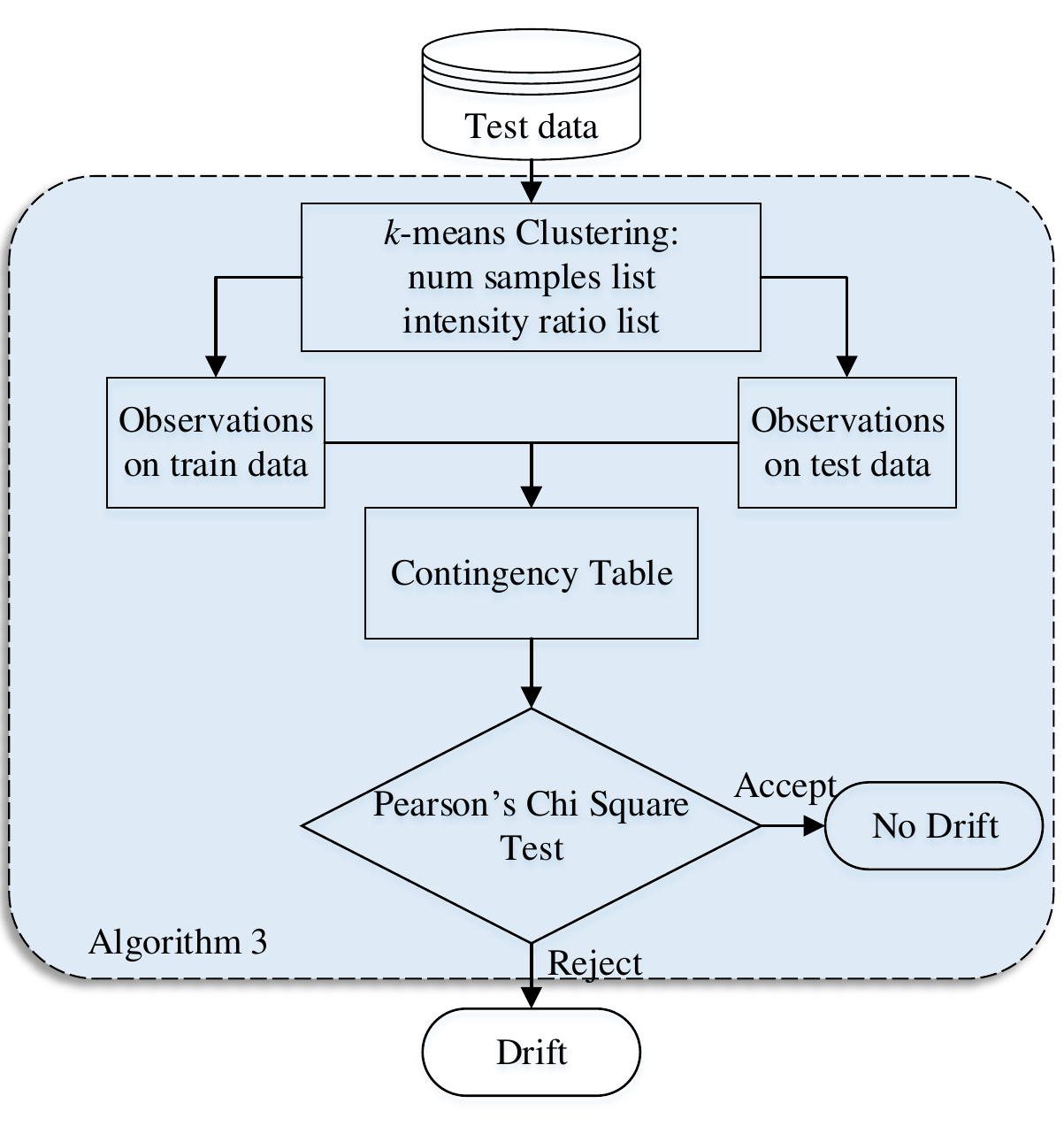}
    \caption{
    The workflow for EI-kMeans detecting drift.
    }
    \label{fig:5_2}
\end{figure}

EI-kMeans considers the clustering property between samples as important when drift occurs. We assume that the distribution change is more likely to occur in a closely-located group of samples than in an arbitrary shape. EI-kMeans space partitioning are cluster-prioritized and are more sensitive to drift within multi-cluster type datasets. The drift detection workflow, in Fig. \ref{fig:5_2}, is simple and fast, once the space partitioning is finished. Based on the output of how the partitions are constructed, the testing samples are clustered into $K$ partitions. The observations in the training and the testing sample sets are vertically stacked to form a contingency table, and the $\chi^2$ test is applied to evaluate the distribution discrepancy between them.

\begin{algorithm}[ht]
    \caption{EI-kMeans Drift Detection}
    \label{alg:3}
    \small
    \SetKwInOut{Input}{input}
    \SetKwInOut{Output}{output}
    \Input{1. Training data set $D_{T_0}$
        \newline 2. Testing set $D_{T_1}$
        \newline 3. EI-kMeans Histogram $(\mathcal{C},V_{coe})$
        }
    \Output{Drift detection results, ($H_0$, $p$-value)}
    \BlankLine
    Find the partition label of $D_{T_0}$ and $D_{T_1}$\;
    Count the frequencies in $D_{T_0}$ and $D_{T_1}$\;
    Build the contingency table\;
    Run Pearson’s chi-square test\;
    return ($H_0$, $p$-value)\;
\end{algorithm}

Algorithm 3 is the drift detection algorithm. It counts the observation frequencies of both the training and testing data, and conducts the $\chi^2$ test. The counting process is implemented using the same steps in Algorithm 2, lines 12-14. If no drift occurs, the observation frequencies of the training data set are stored in the system buffer for the next test. A contingency table is formed for each test by vertically stacking the stored vector and the observation frequencies of the testing data set. The $\chi^2$ test returns a result whether it rejects or accepts the null hypothesis test, denoted as $H_0$. 

The optimized complexity of the 1NN classifier in the EI-kMeans drift detection algorithm is $\mathcal{O}(nlogn)$. The $\chi^2$ test complexity is $\mathcal{O}(K)$. The overall EI-kMeans drift detection runtime complexity is $\mathcal{O}(nK)$. In this algorithm, $n=\max \{ n_{train}, n_{test}\}$

The overall EI-kMeans drift detection algorithm can be summarized into 3 steps.

\begin{itemize}
\item \textbf{Step 1.} Initialize the greedy equal-intensity cluster centroids. 
\item \textbf{Step 2.} Segment the feature space as small clusters. This step is based on k-means clustering, which divides the datasets into a set of individual clusters. This ensures no partition will step across clusters. The number of partitions is continuously reduced if the number of samples in each partition does not satisfy the desired values.
\item \textbf{Step 3.} Detect drift with Pearson’s chi-square test.
\end{itemize}

\section{Experiments and Evaluation}
\label{s:IV}
In this section, we compare the proposed EI-kMeans with other state-of-the-art drift detection algorithms to demonstrate how EI-kMeans performs on the drift detection tasks. The selected histogram-based drift detection algorithms are QuantTree with both $\chi^2$ and total variation statistics which are reported as the best method in their paper \cite{Boracchi:QTree}, kdqTree with $\chi^2$ test \cite{Dasu:kdqTree} and one multivariate two-sample test baseline, known as the multivariate Wald-Wolfowitz test (MWW test) \cite{friedman1979multivariate}. We choose the MWW test as the baseline because it is designed to solve the problem by statistical analysis and its runtime complexity is low enough to perform in a stream learning scenario. To support the reproducible research initiative, the source code of EI-kMeans is available online\footnote{https://github.com/Anjin-Liu/TCYB2019-EIkMeansDriftDetection}

\subsection{A comparison of space partitioning}

\begin{figure*}[ht]
    \centering
    \includegraphics[scale=0.57]{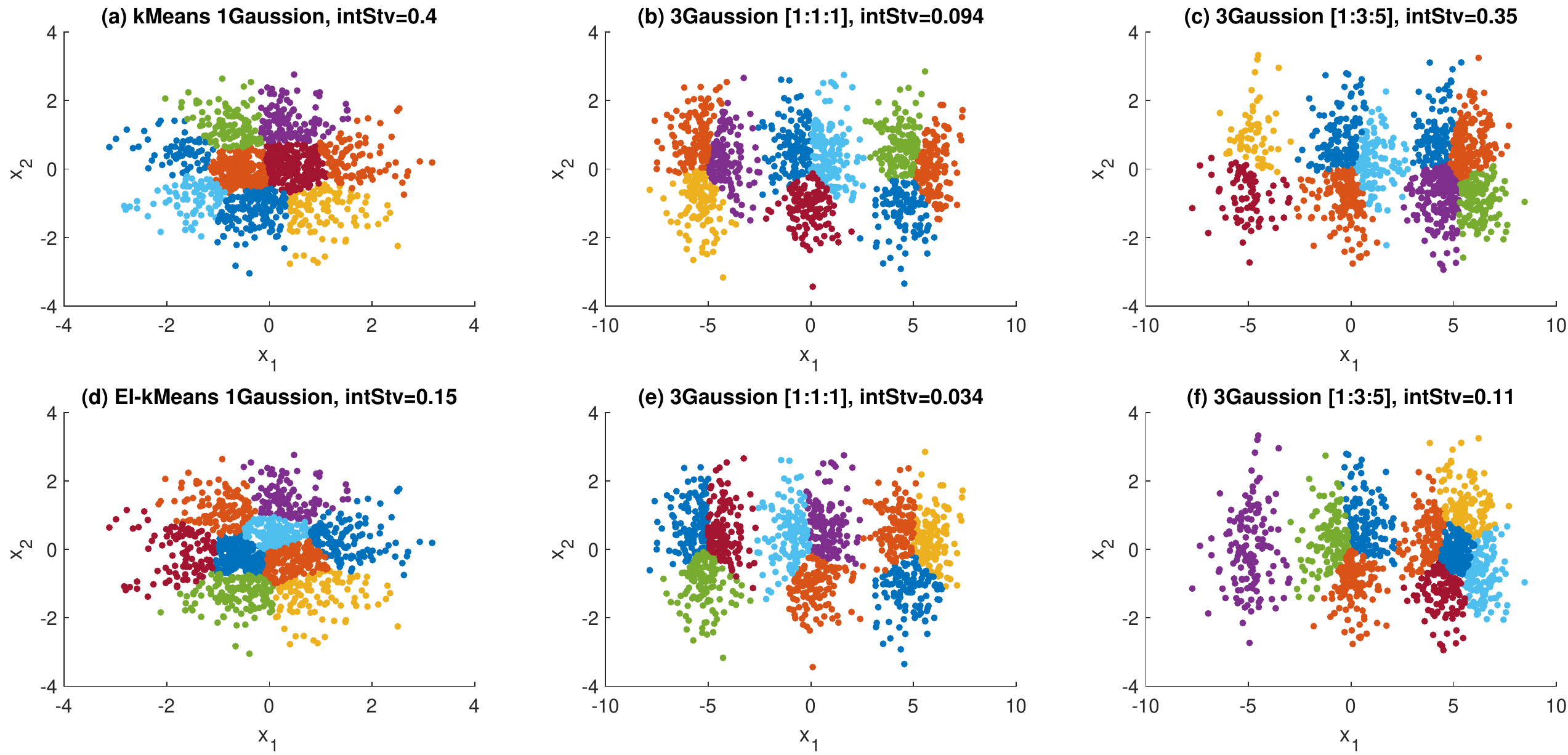}
    \caption{
    A demonstration of EI-kMeans space partitioning compared to kMeans. There is very little difference between panel (a) and (d).  Panel (b) and (e) both have good equal-intensity space partitioning because data is evenly divided between the number of partitions. Panels (c) and (f) show the advantages of EI-kMeans with uneven cluster ratios. The key difference is that kMeans partitions the space according to distance, while EI-kMeans partitions according to intensity and distance. So, as an example, kMeans splits the cluster on the left, while EI-kMeans keeps it whole to ensure the number of partitions for each cluster is the same as its sample ratio. }
    \label{fig:6}
\end{figure*}

\textbf{Experiment 1}. For this experiment, we generated three data sets with different configurations to demonstrate the difference in the space partitioning. The partitioning results are shown in Fig. \ref{fig:6}. The first data set, denoted as 1G, has a Gaussian distribution with a mean of $\mu=[0,0]$, a variance matrix of $\Sigma=
\begin{bmatrix}
1 & 0 \\
0 & 1
\end{bmatrix}$, 
and 1350 data samples. The second data set, denoted as 3G[1:1:1], has three Gaussian distributions with different means: $\mu_1=[-5,0]$, $\mu_2=[0,0]$, $\mu_3=[5,0]$. The variance matrixes are the same, which form three clusters with the same number of data samples in each cluster. The third data set, denoted as 3G[1:3:5], has the same settings as 3G[1:1:1] but with a diverse sample ratio for each cluster, i.e., the cluster with the mean of $\mu_1$ contains 150 data samples; $\mu_2$ has 450 samples; and $\mu_3$ has 750. The number of desired partitions is set as $K=9$. 
 
\textbf{Findings and discussion}: The intStv stands for the standard deviation of the partitions' intensity, which is calculated via Eq. \eqref{eq:intensity}. Low intStv implies that the samples are evenly distributed in each partition. The results shows that no matter what shape of the data set is, EI-kMeans always has a smaller intensity variation than kMeans, which is what we want to achieve.
\begin{table*}[!ht]
\centering
\caption{Configurations of the 2-dimensional synthetic distribution drift data sets. $x$ denotes the feature, $\mu$ denotes the mean vector, $\Sigma$ denotes the variance matrix, and $\delta$ denotes the drift margin.}
\label{t:1}
\begin{tabular}{@{}llll@{}}
\toprule
Data type  & Description                                     & Configurations & Drift margin \\ \midrule
2d-U-mean  & \begin{tabular}[c]{@{}l@{}}Uniform distribution \\ with drift mean\end{tabular}            & $x_1\in[0, 1+\delta]$                         & $\delta=0.06$             \\
2d-1G-mean & \begin{tabular}[c]{@{}l@{}}Gaussian distribution\\ with drift mean\end{tabular}           & $\mu=[0+\delta, 0]$, $\Sigma=\begin{bmatrix}
1 & 0 \\
0 & 1
\end{bmatrix}$                  & $\delta=0.3$             \\
2d-1G-var  & \begin{tabular}[c]{@{}l@{}}Gaussian distribution\\ with drift variance\end{tabular}       & $\mu=[0+\delta, 0]$, $\Sigma=\begin{bmatrix}
1+\delta & 0 \\
0 & 1+\delta
\end{bmatrix}$                                     & $\delta=0.2$             \\
2d-1G-cov  & \begin{tabular}[c]{@{}l@{}}Gaussian distribution\\ with drift covariance\end{tabular}     & $\mu=[0+\delta, 0]$, $\Sigma=\begin{bmatrix}
1 & 0+\delta \\
0+\delta & 1
\end{bmatrix}$                       & $\delta=0.2$             \\
2d-2G-mean & \begin{tabular}[c]{@{}l@{}}2 Gaussian mixture\\ distribution with drift mean\end{tabular} & $\mu_1=[0, 0]$, $\mu_2=[0+\delta, 0]$, $\Sigma_1,\Sigma_2=\begin{bmatrix}
1 & 0 \\
0 & 1
\end{bmatrix}$                        & $\delta=0.4$             \\
2d-4G-mean & \begin{tabular}[c]{@{}l@{}}4 Gaussian mixture\\ distribution with drift mean\end{tabular} & $\mu_1=[0, 0]$, $\mu_2=[5, 0]$,$\mu_3=[0, 5]$,$\mu_4=[5-\delta, 5]$, $\Sigma_1,\Sigma_2,\Sigma_3,\Sigma_4=\begin{bmatrix}
1 & 0 \\
0 & 1
\end{bmatrix}$                       & $\delta=0.8$             \\ \bottomrule
\end{tabular}
\end{table*}

\begin{table*}[!ht]
\centering
\caption{Drift detection results of experiment 2, $n_{train}=2000$, $n_{test}=200$. Each detection algorithm was run 50 times on 250 data sets generated with different random seeds, the average and standard deviation of Type-I error are reported. The underlined values are the Type-I error which exceed the predefined false positive rate, $\alpha=0.05$. 
}
\label{t:2}
\begin{tabular}{@{}lllllll@{}}
\toprule
           & EI-kMeans $\chi^2$ test & kMeans $\chi^2$ test & kdqTree $\chi^2$ test & QuantTree $\chi^2$ stat & QuantTree $TV$ stat & MWW test                                    \\ 
           \midrule
2d-U-mean  & 4.79$\pm$1.57           & 4.02$\pm$1.35        & 5.00$\pm$1.94         & 4.89$\pm$1.48           & 3.77$\pm$1.36       & \underline{10.11$\pm$4.62} \\
2d-1G-mean & 4.90$\pm$1.94           & 3.65$\pm$1.62        & 4.88$\pm$2.26         & 4.72$\pm$1.69           & 3.70$\pm$1.40       & \underline{10.80$\pm$4.60} \\
2d-1G-var  & 4.90$\pm$1.94           & 3.65$\pm$1.62        & 4.88$\pm$2.26         & 4.72$\pm$1.69           & 3.70$\pm$1.40       & \underline{10.80$\pm$4.60} \\
2d-1G-cov  & 4.90$\pm$1.94           & 3.65$\pm$1.62        & 4.88$\pm$2.26         & 4.72$\pm$1.69           & 3.70$\pm$1.40       & \underline{10.80$\pm$4.60} \\
2d-2G-mean & 3.82$\pm$1.70           & 3.14$\pm$1.54        & 4.01$\pm$2.29         & 4.51$\pm$1.97           & 3.04$\pm$1.43       & \underline{10.68$\pm$5.28} \\
2d-4G-mean & 2.31$\pm$1.39           & 2.06$\pm$1.02        & 2.72$\pm$1.62         & 2.66$\pm$1.01           & 2.11$\pm$0.99       & \underline{9.36$\pm$4.65}  \\
\midrule
Average    & 4.27                    & 3.36                 & 4.39                  & 4.37                    & 3.34                & \underline{10.43}       \\ \bottomrule  
\end{tabular}
\end{table*}

\begin{table*}[!ht]
\centering
\caption{The average and standard deviation of Type-II error of experiment 2. The bold values are the lowest Type-II error in the row. The number in the bracket next to the average Type-II error indicates the rank of the average Type-II error. The lower the Type-II error is, the higher rank the result will be.}
\label{t:sd}
\begin{tabular}{@{}lllllll@{}}
\toprule
           & EI-kMeans $\chi^2$ test                   & kMeans $\chi^2$ test & kdqTree $\chi^2$ test                     & QuantTree $\chi^2$ stat & QuantTree $TV$ stat & MWW test                                 \\
           \midrule
2d-U-mean  & 45.00$\pm$10.18                              & 47.87$\pm$9.60       & 47.28$\pm$11.77                           & 40.27$\pm$21.28         & 57.25$\pm$23.72     & \textbf{13.57$\pm$5.64} \\
2d-1G-mean & \textbf{40.84$\pm$11.34} & 58.28$\pm$11.45      & 43.71$\pm$12.07                           & 61.34$\pm$11.39         & 70.33$\pm$7.47      & 83.62$\pm$7.34                           \\
2d-1G-var  & 80.46$\pm$6.02                            & 81.76$\pm$6.78       & \textbf{76.51$\pm$8.95}  & 84.93$\pm$4.43          & 89.85$\pm$3.22      & 83.72$\pm$7.38                           \\
2d-1G-cov  & \textbf{80.84$\pm$}5.51  & 87.47$\pm$4.58       & 83.39$\pm$5.82                            & 90.72$\pm$5.39          & 92.45$\pm$4.57      & 87.20$\pm$6.18                           \\
2d-2G-mean & \textbf{57.43$\pm$}10.80 & 66.78$\pm$9.66       & 59.78$\pm$11.41                           & 70.57$\pm$10.42         & 83.74$\pm$5.20      & 83.22$\pm$6.90                           \\
2d-4G-mean & 46.42$\pm$13.37                           & 53.58$\pm$12.70      & \textbf{45.88$\pm$}13.27 & 77.21$\pm$12.17         & 84.32$\pm$9.08      & 80.70$\pm$7.77                           \\ \midrule
Average    & \textbf{58.50} (1)       & 65.96 (3)            & 59.43 (2)                                 & 70.84 (4)               & 79.66 (6)           & 72.00 (5)                \\\bottomrule               
\end{tabular}
\end{table*}

\begin{table*}[!ht]
\centering
\caption{Drift detection results of EI-kMeans for experiment 2 with different training batch size.}
\label{t:2_diff}
\begin{tabular}{@{}lllllllll@{}}
\toprule
\multirow{2}{*}{} & \multicolumn{2}{l}{$n_{train}=2000$} & \multicolumn{2}{l}{$n_{train}=3000$} & \multicolumn{2}{l}{$n_{train}=4000$} & \multicolumn{2}{l}{$n_{train}=5000$} \\ \cmidrule(l){2-9} 
                  & Type-I         & Type-II        & Type-I           & Type-II        & Type-I           & Type-II        & Type-I         & Type-II        \\ \midrule
2d-U-mean         & 4.79           & 45.00          & 4.83             & 43.32          & 4.84             & 41.57          & 4.78           & \textbf{40.86} \\
2d-1G-mean        & 4.90           & 40.84          & \underline{5.06} & 38.39          & \underline{5.17} & 37.55          & 5.00           & \textbf{36.77} \\
2d-1G-var         & 4.90           & 80.46          & \underline{5.06} & \textbf{79.57} & \underline{5.17} & 79.85          & 5.00           & \textbf{79.57} \\
2d-1G-cov         & 4.90           & 80.84          & \underline{5.06} & \textbf{80.10} & \underline{5.17} & 80.39          & 5.00           & 80.72          \\
2d-2G-mean        & 3.82           & 57.43          & 3.73             & 55.69          & 3.68             & \textbf{54.74} & 3.64           & 55.52          \\
2d-4G-mean        & 2.31           & 46.42          & 2.13             & 45.55          & 2.06             & 45.18          & 1.98           & \textbf{44.43} \\ \midrule
Average           & 4.27           & 58.50 (4)      & 4.31             & 57.10 (3)      & 4.35             & 56.55 (2)      & 4.23           & \textbf{56.31} (1)      \\ \bottomrule
\end{tabular}
\end{table*}
\subsection{Drift detection accuracy with synthetic data sets}

\begin{figure*}[ht]
    \centering
    \includegraphics[scale=0.58]{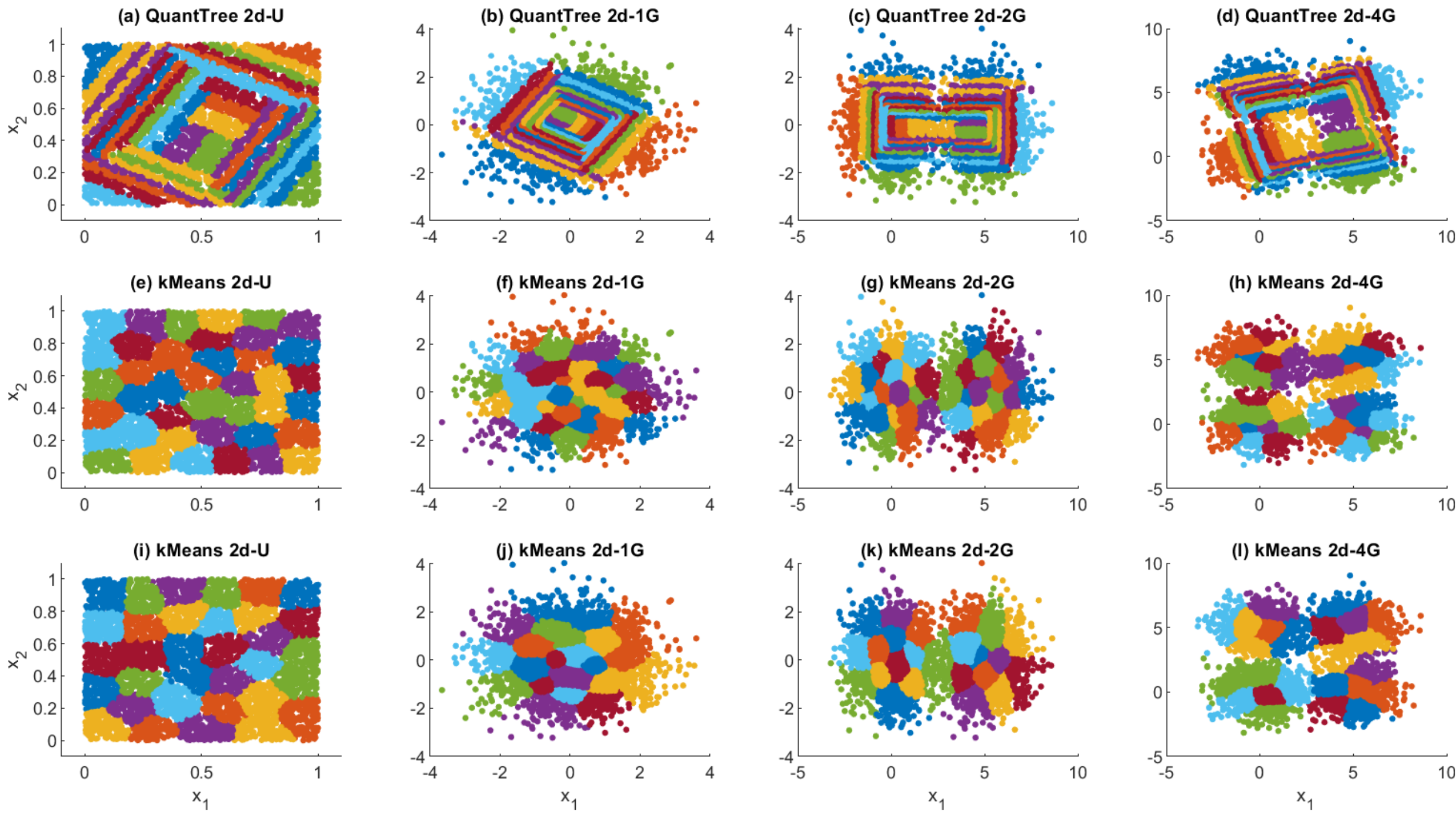}
    \caption{
    Space partitioning results of experiment 2. As shown in sub-figure (a)-(d) the partitions of QuantTree are rectangle-based and cross multiple clusters, which may jeopardise its capability to drift detection.  Sub-figures (e)-(h) are the partitions created by kdqTree algorithm. All partitions have similar size no matter how many samples are inside. This could be dangerous on high density variation samples, because some sparse regions may not have enough samples to estimate the density. For the sub-figure (i)-(l), kMeans shows compact partitions. However, some partitions are too small to include enough samples, which may jeopardise its drift detection sensitiveness. Sub-figure (m)-(p) are the partitions of EI-kMeans, which is very similar to kMeans. This is anticipated since we used the same cluster algorithm. The main difference is that EI-kMeans ensures each partition could have a reasonable large size, so that the number of sample in each partition are evenly distributed. 
    }
    \label{fig:2_1_space}
\end{figure*}

\textbf{Experiment 2}:  In this experiment, we generated six 2-dimensional data sets to evaluate the power of EI-kMeans to detect drift. We compared EI-kMeans with the state-of-the-arts QuantTree, kdqTree and k-means space partition plus $\chi^2$ test. The training set contained 2000 training samples, and the testing set contained 200 samples. For each data type, we generated 250 stationary testing sets and 250 drift testing sets and evaluated both Type I and Type II errors. Type I errors are rejections of a true null hypothesis (also known as a "false positive"). A Type II error is the false null hypothesis rates (a "false negative"). The Type-I and Type-II errors are the most common evaluation metric for distribution change detection. To evaluate the stability, we run the test 50 times and recorded the mean and standard deviation. Table \ref{t:1} presents the data set configurations, and Table \ref{t:2} shows the mean of the drift detection results. Table \ref{t:sd} shows the standard deviation. To evaluate the influence of training batch size, we changed the training set size to 3000, 4000 and 5000. The detection results are shown in Table \ref{t:2_diff}. Fig. \ref{fig:2_1_space}. shows the space partitioning results of each algorithm.

\textbf{Findings and discussion}: The results shows that all drift detection algorithms outperformed the base-line multivariate two-sample test, MWW test. The results demonstrate that EI-kMeans with $\chi^2$ test has the average Type-I error below $\alpha=0.05$ as well as the lowest average Type-II error. Comparing to the kMeans-based space partition the improvement of EI-kMeans space partitioning is significant, which raised the rank from (3) to (1). The kdqTree space partition with $\chi^2$ test performed well in this experiment, and had shown no significant disadvantages compared to others. The QuantTree space partitioning with $\chi^2$ and $TV$ statistics are not performing well in general, because the partitioning strategy is not designed for multi-cluster data sets. As we can see, the Type-II error of QuantTree $\chi^2$ stat is very close to the kMeans $\chi^2$ test on the 2d-U-mean, 2d-1G-mean, 2d-1G-var and 2d-1G-cov data sets, which are all single cluster type data sets. Average performance dropped significantly on the multi-cluster data sets 2d-2G-mean and 2d-4G-mean. Based on these results, we conclude that the design of a histogram scheme makes a significant contribution to the drift detection accuracy in different data distribution which is a nontrivial problem. Regarding to the batch size, as we use Pearson's chi-square test as the drift detection hypothesis test, the drift threshold of the test statistics is determined by the Chi-square distribution with a given significant level. A sample set with a sufficiently large size is assumed. If a chi-squared test is conducted on a sample with a small size, the chi-squared test will yield an inaccurate inference, which might end up committing a Type II error. As we can see, the Type-II error increases as the training size decrease.

\begin{figure}[ht]
    \centering
    \includegraphics[scale=0.55]{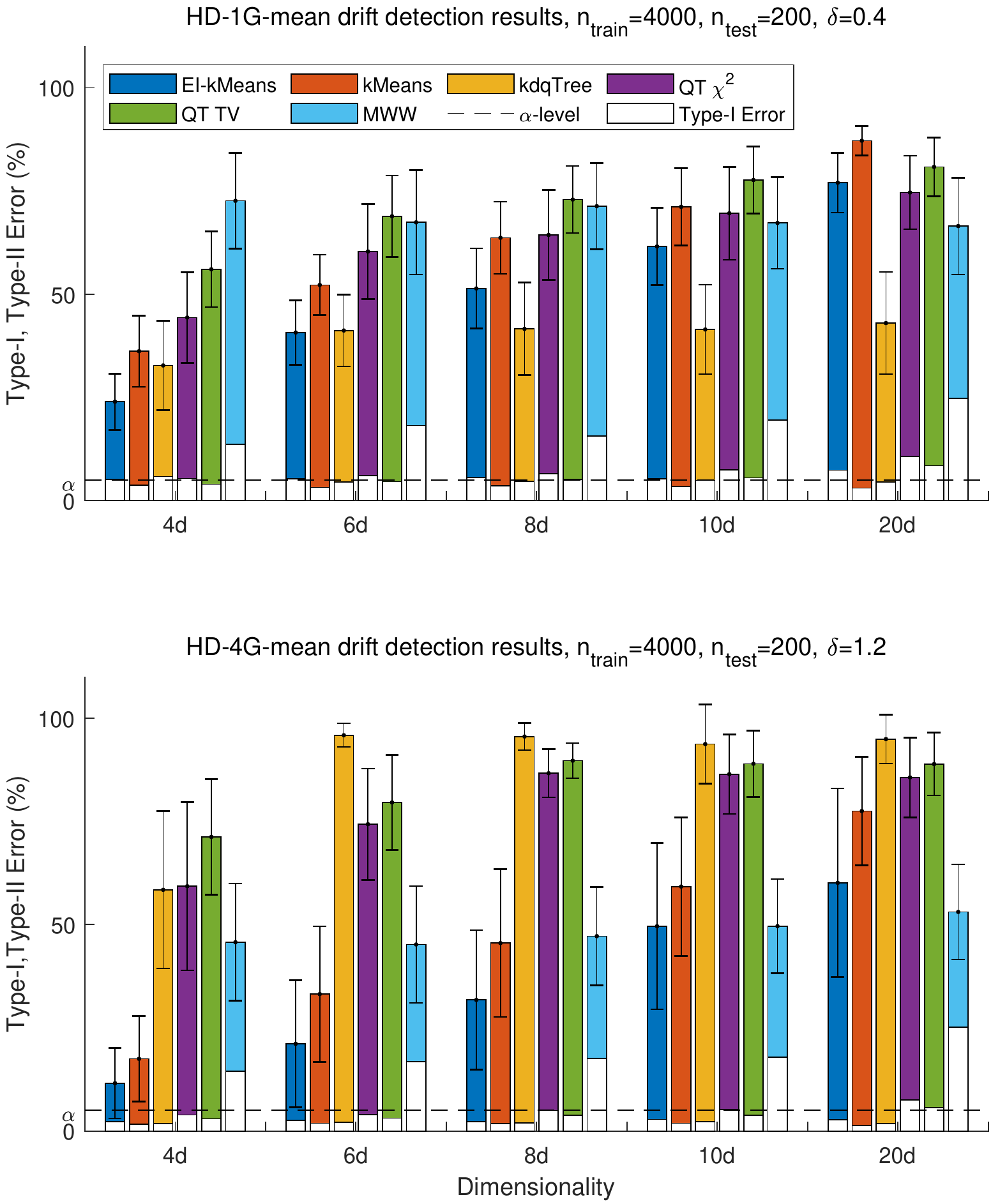}
    \caption{
    The result of experiment 3. The bar chars indicates the average Type-I, Type-II errors in percentage for each algorithm. The $x$-axis is the dimenionality of the data sets. The dash line is the predefined $\alpha$-level.
    }
    \label{fig:2_2}
\end{figure}

\textbf{Experiment 3}: To evaluate the proposed algorithm on high dimensional data, we expand the 2d-1G-mean and 2d-4G-mean data sets to 4, 6, 8, 10 and 20 dimensions by adding normal distributed data. For example, in the 4d-1G-mean data set, the first two features are the same as the 2d-1G-mean but, for the third and fourth features, they are generated by normal distribution $\mathcal{N}(0,1)$ with covariance equal to 0. Since increasing unrelated dimensions will reduce the drift sensitiveness, we increased the drift margin for HD-1G-mean to 0.5, for HD-4G-mean to 1.0, and doubled the training data. The results are given in Fig. \ref{fig:2_2}.

\textbf{Findings and discussion}: In Fig. \ref{fig:2_2}, the Type-II errors increased as the non-drift dimension increased. Most algorithms preserved a low Type-I error, except the MWW test. Although MWW test has the lowest Type-II error on the HD-4G-mean drift data sets, its Type-I error is above the desired $\alpha$-level threshold. The kdqTree with $\chi^2$ test has the best performance on the HD-1G-mean data sets, but it turns to powerless on the HD-4G-mean data sets. We consider this is because kdqTree does have a effective method to control the number of samples in each partition. Directly applying $\chi^2$ test on the kdqTree partitions is risky. In this experiment, EI-kMeans outperforms others in most cases and has its false positive rate below the predefined threshold $\alpha=0.05$, which indicates that it is stable on high dimensional data.

\subsection{Drift detection accuracy with real-world data sets}

\textbf{Experiment 4.} Drift detection on real-world data sets.
For this experiment, we created 8 train-test drift detection tests from 5 real-world data applications. For each test, we generated one training data set and 500 testing data sets. Among these testing data sets, half were drawn from the same distribution, the other half were drawn from a different distribution. Again, the results were evaluated in terms of Type-I and Type-II errors. The characteristics of these data sets are summarized in Table \ref{t:6}.

\textbf{HIGGS Bosons and Background Data Set}. The objective of this data set is to distinguish the signatures of the processes that produce Higgs boson particles from those background processes that do not. Four low-level indicators of the azimuthal angular momenta for four particle jets were selected as features, which means the distributions were $\mathbb{R}^4$. The jet momenta distributions of the background processes are denoted as Back, and the processes that produce Higgs bosons are denoted as Higgs. The total sample size of mixed Backs and Higgs is $1.1\times 10^7$. We randomly selected 2000 samples without replacement from each distribution as the training data.1000 samples were used as the testing set. There were three types of data integration: Back-Back where both sample sets were drawn from Back; Higgs-Higgs, where both sample sets were drawn from Higga; and Higgs-Back, where one sample set was drawn from Higgs and the other from Back.

\textbf{MiniBooNe Particle Identification Data Set}. This data set contains 36,499 signal events and 93,565 background events. Each event has 50 particle ID variables. The drift detection task is to distinguish between signal events and background events. The sample size of the training set size was 2000, and 500 for the testing set.

\textbf{Arabic Digit Mixture Data Set}. This data set contains audio features of 88 people (44 females and 44 males) pronouncing Arabic digits between 0 and 9. We applied the same configuration as Denis et al. \cite{FastKSTest}. The data set was originally i.i.d. and contained a time series for 13 different attributes. The revised configuration has 26 attributes instead of 13 time series with a replacement mean and standard deviation for each time series. Mixture distributions were generated by grouping female and male labels. Mixture distribution $\mathcal{A}$ contained randomly selected samples of both males and females, with male labels from 0 to 4 and female labels from 5 to 9. Mixture distribution $\mathcal{B}$ reversed the labels at 9, i.e., drawing the samples of $\mathrm{ID}_9$ with label male. We configured the data set this way to create multiple clusters, where the pronunciation of each digit formed a cluster. The configuration is summarized in Table \ref{t:3}. The training set size was 2000, and the testing set size was 500.
\begin{table*}[!ht]
\centering
\caption{Data generation structure of the Arabic digit mixture data set (F: female, M: male).}
\label{t:3}
\begin{tabular}{@{}lllllllllll@{}}
\toprule
               & $\mathrm{ID}_0$ & $\mathrm{ID}_1$ & $\mathrm{ID}_2$ & $\mathrm{ID}_3$ & $\mathrm{ID}_4$ & $\mathrm{ID}_5$ & $\mathrm{ID}_6$ & $\mathrm{ID}_7$ & $\mathrm{ID}_8$ & $\mathrm{ID}_9$ \\ \midrule
$\mathcal{A}$ & M & M & M & M & F & F & F & F & F & F \\
$\mathcal{B}$ & M & M & M & M & F & F & F & F & F & M \\ \bottomrule
\end{tabular}
\end{table*}

\textbf{Localization Mixture Data Set}. The localization data set contains data from a sensor carried by 5 different people (A, B, C, D, E). The original data has 11 different movements with imbalanced samples. To use this data set for drift detection, we selected the top three movements with the most samples, 'lying', 'walking' and 'sitting'. To simulate multiple clusters with drift, we grouped samples from different people together at different percentages to result in varied data distributions. The training set size was 3000, and 600 for the testing set. The sample proportion of each people is summarised in Table \ref{t:4}.
\begin{table}[!ht]
\centering
\caption{Data generation structure of the localization mixture data set. The values are the proportion of samples drawn for each person.}
\label{t:4}
\begin{tabular}{@{}llllll@{}}
\toprule
               & $\mathrm{People}_A$   & $\mathrm{People}_B$   & $\mathrm{People}_C$   & $\mathrm{People}_D$   & $\mathrm{People}_E$   \\ \midrule
$\mathcal{A}$ & 0.0 & 0.2 & 0.2 & 0.2 & 0.4 \\
$\mathcal{B}$ & 0.4 & 0.4 & 0.2 & 0.0 & 0.0 \\ \bottomrule
\end{tabular}
\end{table}

\textbf{Insects Mixture Data Set}. This data set contains features from a laser sensor. The task is to distinguish between 5 possible specimens of flying insects that pass through a laser in a controlled environment (Flies, Aedes, Tarsalis, Quinx, and Fruit). A preliminary analysis showed no drift in the feature space. However, the class distributions gradually change over time. To simulate drift in multiple clusters, we selected the samples from different insects and grouped them together at different percentages. Thus, the data distribution may vary. The training set size was 2000, and 500 for the testing set size. The sample proportion of each specimens is summarized in Table \ref{t:5}.
\begin{table}[!ht]
\centering
\caption{Data generation structure of the insect mixture data set. The values are the proportion of samples drawn for each insect type.}
\label{t:5}
\begin{tabular}{@{}llllll@{}}
\toprule
               & Flies & Aedes & Tarsalis & Quinx & Fruit \\ \midrule
$\mathcal{A}$ & 0.2   & 0.2   & 0.2      & 0.2   & 0.2   \\
$\mathcal{B}$ & 0.14  & 0.14  & 0.2      & 0.2   & 0.32  \\ \bottomrule
\end{tabular}
\end{table}

\begin{table}[!ht]
\centering
\caption{The characteristics of the data sets.}
\label{t:6}
\begin{tabular}{@{}lllll@{}}
\toprule
Data set ID & Data set name    & \# Features & \# Training & \# Testing  \\ \midrule
Real-I      & Higgs-Back       & 4           & 2000                & 1000               \\
Real-II     & Back-Higgs       & 4           & 2000                & 1000               \\
Real-III    & Sign-Back        & 50          & 2000                & 500                \\
Real-IV     & Back-Sign        & 50          & 2000                & 500                \\
Real-V      & Arabic $\mathcal{A}$-$\mathcal{B}$ & 26          & 2000                & 500                \\
Real-VI     & Arabic $\mathcal{B}$-$\mathcal{A}$ & 26          & 2000                & 500                \\
Real-VII    & Localization     & 3           & 3000                & 600                \\
Real-VIII   & Insect           & 49          & 2000                & 500                \\ \bottomrule
\end{tabular}
\end{table}

\begin{table*}[!ht]
\centering
\caption{Drift detection results with real-world data sets, Type-I Error (\%). Each detection algorithm was run 50 times on 250 data sets generated with different random seeds, and the average Type-I error and the standard deviation are reported. The underlined results exceed the predefined false positive rate, $\alpha=0.05$.}
\label{t:7_tI}
\begin{tabular}{@{}lllllll@{}}
\toprule
          & EI-kMeans $\chi^2$ test                    & kMeans $\chi^2$ test & kdqTree $\chi^2$ test                      & QuantTree $\chi^2$ stat                    & QuantTree $TV$ stat                        & MWW test                                     \\ \midrule
Real-I    & \underline{6.34$\pm$5.97} & 3.50$\pm$2.72        & 4.80$\pm$4.18                              & \underline{6.70$\pm$5.73} & \underline{5.77$\pm$4.85} & \underline{7.14$\pm$5.94}   \\
Real-II   & \underline{9.09$\pm$8.13} & 3.44$\pm$3.37        & 4.06$\pm$3.24                              & \underline{5.26$\pm$3.15} & 4.34$\pm$2.71                              & \underline{6.54$\pm$5.72}   \\
Real-III  & 4.67$\pm$2.74                              & 1.8$\pm$1.35         & 3.12$\pm$3.59                              & 3.83$\pm$1.9                               & 3.69$\pm$1.80                              & 3.14$\pm$3.60                                \\
Real-IV   & 4.06$\pm$2.03                              & 1.94$\pm$1.52        & 2.94$\pm$2.50                              & 4.91$\pm$2.68                              & 4.59$\pm$2.39                              & 3.26$\pm$3.21                                \\
Real-V    & 4.40$\pm$4.14                              & 3.96$\pm$3.97        & \underline{5.37$\pm$5.33} & 4.42$\pm$4.11                              & 4.30$\pm$3.96                              & 1.46$\pm$3.05                                \\
Real-VI   & 4.45$\pm$3.94                              & 2.96$\pm$2.35        & 4.81$\pm$4.20                              & \underline{5.34$\pm$6.45} & 4.87$\pm$5.16                              & 1.33$\pm$1.84                                \\
Real-VII  & 0.00$\pm$0.00                              & 0.00$\pm$0.00        & 0.00$\pm$0.00                              & 2.00$\pm$14.14                             & 2.00$\pm$14.14                             & \underline{10.00$\pm$30.30} \\
Real-VIII & 2.81$\pm$2.59                              & 1.58$\pm$1.60        & 3.42$\pm$7.90                              & \underline{5.66$\pm$4.96} & 4.97$\pm$4.24                              & \underline{10.45$\pm$9.39}  \\
\midrule
Average   & 4.48                                       & 2.40                 & 3.57                                       & 4.77                                       & 4.32                                       & \underline{5.42}            \\ \bottomrule
\end{tabular}
\end{table*}

\begin{table*}[!ht]
\centering
\caption{Drift detection results with real-world data sets, Type-II Error (\%). The bold results are the lowest Type-II error on this data set.}
\label{t:7_tII}
\begin{tabular}{@{}lllllll@{}}
\toprule
 & EI-kMeans $\chi^2$ test & kMeans $\chi^2$ test                      & kdqTree $\chi^2$ test                   & QuantTree $\chi^2$ stat                   & QuantTree $TV$ stat                     & MWW test                                         \\ \midrule
Real-I                  & 87.16$\pm$9.69                            & 85.34$\pm$9.46                          & \textbf{78.61$\pm$11.69} & 80.00$\pm$13.10                         & 82.76$\pm$10.84                         & 89.45$\pm$7.85                            \\
Real-II                 & 74.02$\pm$17.21                           & 92.73$\pm$5.77                          & 77.71$\pm$13.48                           & 79.66$\pm$11.62                         & 84.46$\pm$8.80                          & \textbf{69.78$\pm$19.22} \\
Real-III                & \textbf{0.00$\pm$0.00}   & \textbf{0.00$\pm$0.00} & 25.94$\pm$43.94                           & \textbf{0.00$\pm$0.00} & \textbf{0.00$\pm$0.00} & \textbf{0.00$\pm$0.00}   \\
Real-IV                 & \textbf{0.00$\pm$0.00}   & \textbf{0.00$\pm$0.00} & 3.90$\pm$12.02                            & \textbf{0.00$\pm$0.00} & \textbf{0.00$\pm$0.00} & \textbf{0.00$\pm$0.00}   \\
Real-V                  & \textbf{12.92$\pm$12.61} & 18.94$\pm$13.18                         & 16.78$\pm$10.90                           & 66.68$\pm$17.64                         & 84.71$\pm$9.06                          & 98.15$\pm$1.99                            \\
Real-VI                 & 12.10$\pm$12.92                           & \textbf{4.26$\pm$5.38} & 17.15$\pm$10.51                           & 92.03$\pm$7.82                          & 93.98$\pm$6.51                          & 98.82$\pm$1.23                            \\
Real-VII                & 62.00$\pm$49.03                           & 36.00$\pm$48.49                         & 58.00$\pm$49.86                           & 62.00$\pm$49.03                         & 64.00$\pm$48.49                         & \textbf{2.00$\pm$14.14}  \\
Real-VIII               & \textbf{12.37$\pm$10.98} & 32.62$\pm$17.45                         & 30.22$\pm$17.35                           & 77.32$\pm$14.40                         & 80.90$\pm$11.92                         & 68.66$\pm$18.43                           \\
\midrule
Average                 & \textbf{32.57} (1)       & 33.74 (2)                               & 38.54 (3)                                 & 57.21 (5)                               & 61.35 (6)                               & 53.36 (4)                                 \\ \bottomrule
\end{tabular}
\end{table*}

\textbf{Findings and discussion}: The average drift detection accuracy is shown in Table \ref{t:7_tI} \ref{t:7_tII}, and their standard deviation. The results show that all tested methods returned an average false positive rate below the $\alpha=0.05$ except MWW test. EI-kMeans had the lowest average Type-II error of 32.57\%, which is 1.17\% lower than the next best performance by k-means with a $\chi^2$ test. However, EI-kMeans improved drift detection power comes at the cost of an increased false positive rate, and sometimes at over the predefined thresholds. This result conforms to our expectation since EI-kMeans places more restrictive constraints on the number of samples in each partition to meet the requirements of $\chi^2$ test. With a small sample set, the $\chi^2$ test will yield an inaccurate inference and is prone to Type II errors. Notably, however, while the true positive detection accuracy may be impaired, the false alarm rate does not surpass the predefined threshold $\alpha$. 

Across the Real-I to Real-VII data sets, the cluster-based algorithms performed just as well as the tree-based algorithms. However, the QuantTree algorithms completely lost its power to detect drift with the Real-VIII Insect mixture cluster-based data set, while EI-kMeans showed the best performance.


\section{Conclusions and Future Work}
\label{s:V}

In this paper, we proposed a novel space partitioning algorithm, called EI-kMeans, for drift detection on multi-cluster data sets. EI-kMeans is a modified k-means algorithm to search for the best centroids to create partitions. The distances between samples and centroids are amply-shrink based on the cluster intensity ratios. The proposed algorithm detects concept drift from a data distribution perspective. Similar to most distribution-based drift detection algorithms, with a supervised learning setting, it will trigger drift alarm if there is a real or virtual drift but it may not be able to distinguish the drift types. The results of our experiments show the power of EI-kMeans to detect drift with multi-cluster type data sets and proved that histogram design is critical to drift detection accuracy. The results also show that uniform space partitioning may not always outperform other schemes – the performance of the space partition algorithm is data-dependent.

The version of EI-kMeans considered in this paper is designed for a Pearson’s chi-square test, but different hypothesis tests may require different methods of histogram construction. This is something we intend to explore in future work. In addition, concept drift detection is only one aspect of learning in a dynamic stream. How to design a tailored drift adaptation algorithm that leverages the drift detection result to achieve better performance in stream learning is our next target.

\section*{Acknowledgment}
The work presented in this paper was supported by the Australian Research Council (ARC) under Discovery Project DP190101733. We acknowledge the support of NVIDIA Corporation with the donation of GPU used for this research.

\ifCLASSOPTIONcaptionsoff
  \newpage
\fi



\bibliographystyle{jabbrv_ieeetr}
\bibliography{bib/reference}
%



%

\begin{IEEEbiography}[{\includegraphics[width=1in,height=1.25in, clip]{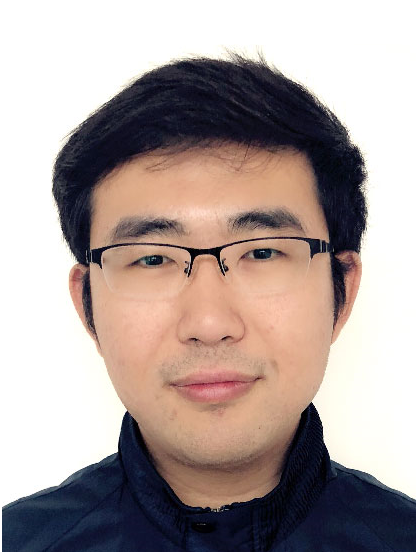}}]{Anjin Liu} (M'17)
is a Postdoctoral Research Associate in the A/DRsch Centre for Artificial Intelligence, Faculty of Engineering and Information Technology, University of Technology Sydney, Australia. He received the BIT degree (Honour) at the University of Sydney in 2012. His research interests include concept drift detection, adaptive data stream learning, multi-stream learning, machine learning and big data analytics.
\end{IEEEbiography}
\begin{IEEEbiography}[{\includegraphics[width=1in,height=1.25in,clip,keepaspectratio]{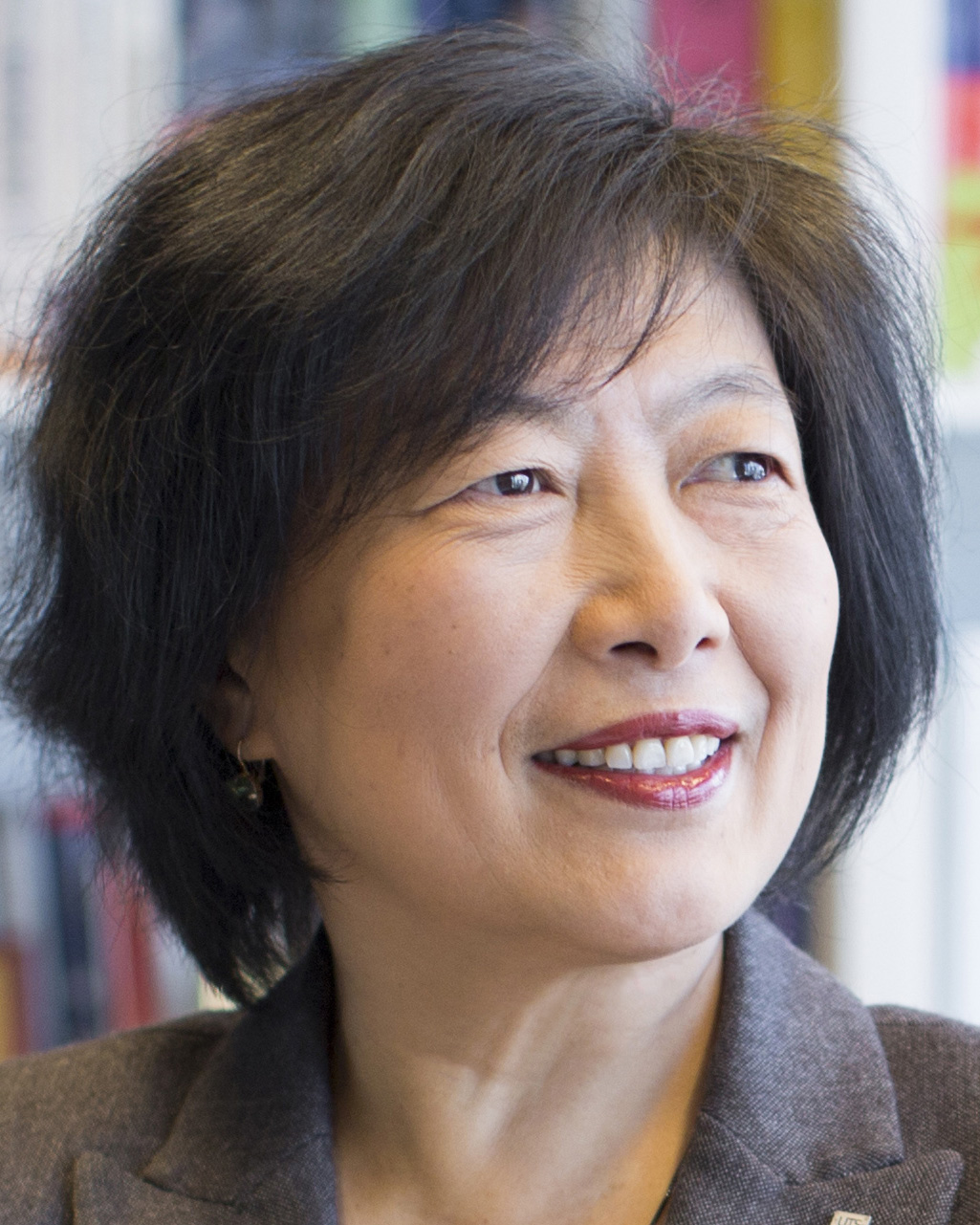}}]{Jie Lu}
(F’18) is a Distinguished Professor and the Director of the Centre for Artificial Intelligence at the University of Technology Sydney, Australia. She received her PhD degree from Curtin University of Technology, Australia, in 2000. 
Her main research interests arein the areas of fuzzy transfer learning, concept drift, decision support systems, and recommender systems. She is an IEEE fellow, IFSA fellow and Australian Laureate fellow. She has published six research books and over 450 papers in refereed journals and conference proceedings; has won over 20 ARC Laureate, ARC Discovery Projects, government and industry projects. She serves as Editor-In-Chief for Knowledge-Based Systems (Elsevier) and Editor-In-Chief for International journal of computational intelligence systems. She has delivered over 25 keynote speeches at international conferences and chaired 15 international conferences. She has received various awards such as the UTS Medal for Research and Teaching Integration (2010), the UTS Medal for Research Excellence (2019), the Computer Journal Wilkes Award (2018), the IEEE Transactions on Fuzzy Systems Outstanding Paper Award (2019), and the Australian Most Innovative Engineer Award (2019).
\end{IEEEbiography}
\begin{IEEEbiography}[{\includegraphics[width=1in,height=1.25in, clip]{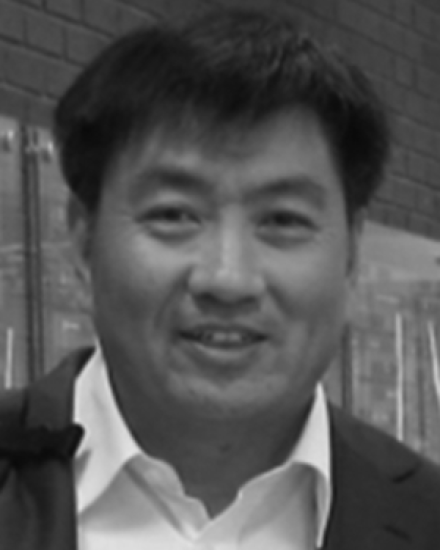}}]{Guangquan Zhang}
is an Associate Professor and Director of the Decision Systems and e-Service Intelligent (DeSI) Research Laboratory at the University of Technology Sydney, Australia. He received the Ph.D. degree in applied mathematics from Curtin University of Technology, Australia, in 2001.
His research interests include fuzzy machine learning, fuzzy optimization, and machine learning. He has authored five monographs, five textbooks, and 460 papers including 220 refereed international journal papers
Dr. Zhang has won seven Australian Research Council (ARC) Discovery Projects grants and many other research grants. He was awarded an ARC QEII fellowship in 2005.
He has served as a member of the editorial boards of several international journals, as a guest editor of eight special issues for IEEE transactions and other international journals, and co-chaired several international conferences and workshops in the area of fuzzy decision-making and knowledge engineering.
\end{IEEEbiography}




\end{document}